\documentclass[sn-apa]{sn-jnl}

\usepackage{graphicx}

\usepackage{hyperref, xcolor}

\usepackage{algorithm}
\usepackage{amsmath}
\usepackage{amssymb}
\usepackage{wrapfig}
\usepackage{multirow}
\usepackage{booktabs}
\usepackage{multirow}
\usepackage{placeins}
\usepackage{cleveref}
\crefformat{footnote}{#2\footnotemark[#1]#3}

\usepackage{phonetic}
\usepackage{subfig}
\usepackage{rotating}
\usepackage{booktabs}
\usepackage{float}
\usepackage{tipa}

\jyear{2023}%

\theoremstyle{thmstyleone}%
\theoremstyle{thmstyletwo}%

\theoremstyle{thmstylethree}%

\begin{document}

\title[CoMadOut - A Robust Outlier Detection Algorithm based on CoMAD]{CoMadOut - A Robust Outlier Detection Algorithm based on CoMAD}


\author*[1]{\fnm{Andreas} \sur{Lohrer}}\email{alo@informatik.uni-kiel.de} 

\author[1]{\fnm{Daniyal} \sur{Kazempour}}\email{dka@informatik.uni-kiel.de} 

\author[1]{\fnm{Maximilian} \sur{Hünemörder}}\email{mah@informatik.uni-kiel.de} 

\author[1]{\fnm{Peer} \sur{Kröger}}\email{pkr@informatik.uni-kiel.de} 


\affil[1]{\orgname{Christian-Albrechts-Universität zu Kiel}, \orgaddress{\street{Christian-Albrechts-Platz 4}, \city{Kiel}, \postcode{24118}, \state{Schleswig-Holstein}, \country{Germany}}}

\abstract{
Unsupervised learning methods are well established in the area of anomaly detection and achieve state of the art performances on outlier datasets. Outliers play a significant role, since they bear the potential to distort the predictions of a machine learning algorithm on a given dataset. Especially among PCA-based methods, outliers have an additional destructive potential regarding the result: they may not only distort the orientation and translation of the principal components, they also make it more complicated to detect outliers. To address this problem, we propose the robust outlier detection algorithm CoMadOut, which satisfies two required properties: (1) being robust towards outliers and (2) detecting them. Our CoMadOut outlier detection variants using comedian PCA define, dependent on its variant, an inlier region with a robust noise margin by measures of in-distribution (variant CMO) and optimized scores by measures of out-of-distribution (variants CMO*), e.g. kurtosis-weighting by CMO+k. These measures allow distribution based outlier scoring for each principal component, and thus, an appropriate alignment of the degree of outlierness between normal and abnormal instances. Experiments comparing CoMadOut with traditional, deep and other comparable robust outlier detection methods showed that the performance of the introduced CoMadOut approach is competitive to well established methods related to average precision (AP), area under the precision recall curve (AUPRC) and area under the receiver operating characteristic (AUROC) curve. In summary our approach can be seen as a robust alternative for outlier detection tasks.
}



\maketitle

\section{Introduction}
\label{sect:intro}
Anomaly Detection, one of the major fields of unsupervised machine learning, is an integral part of many domains uncovering the deviations of their data generating processes and thus supporting domain experts in their daily work by identifying rare patterns. However, real world datasets are often highly imbalanced due to the rare occurrence of outliers. Therefore, for predictive methods, the existence of outliers represents obstacle and opportunity at the same time. Depending on the machine learning method, it can be considered as an obstacle, because they may distort the precision of the predictions. On the other hand, methods can utilize outliers as an opportunity, since they may reveal irregular or abnormal patterns and therefore interesting insights. From the aforementioned two aspects, one can derive two properties that an anomaly detection algorithm needs to satisfy, namely (1) to be resilient towards outlying data instances, while at the same time (2) being capable to detect them. A preliminary step of many outlier detection approaches (cf. Section~\ref{sect:relwork}) is dimensionality reduction. This part is often realized by PCA\cite{JolliffePCA}, a technique providing the directions of highest variances within the data. This is achieved by calculating the eigenvectors of the covariance matrix (directions) and the corresponding eigenvalues (variances). Such eigenpairs allow a transformation from the original data to principal components in lower dimensional subspace, and therewith the task of dimensionality reduction and others. However, the usage of the mean dependent covariance matrix makes the standard PCA potentially susceptible towards outliers, which has lead to the creation of robust PCA methods, minimizing the influence of outliers.\cite{DBLP:journals/jacm/CandesLMW11} In order to achieve resilience or robustness for PCA-based algorithms, we need to ensure that principal components and corresponding eigenvalues are barely or not at all influenced by instances located out of distribution. In particular this means that robustness is achieved if abnormal instances do not skew the orientation and translation of the eigenvectors and do not lead to an increase or decrease of the eigenvalues.

In this work we introduce CoMadOut, an unsupervised outlier detection method, which shows robustness among other techniques due to optimized scoring and outlier resistant PCA\cite{outlcomedianapproach, DBLP:conf/sisap/KazempourH019}. While both works demonstrate that co-median PCA is potentially robust towards anomalies, it lacks the ability to detect abnormal instances based on comedian matrix in its non-positive-semi-definite form. Therefore, with this paper, we propose a competitive outlier detection approach called CoMadOut which is capable to detect, score and predict outliers. 

Since there already exists a plethora of outlier detection methods, we investigate with this work in the performance of CoMadOut against a comprehensive selection of state-of-the-art techniques~(cf. section~\ref{sect:exp}) covering both traditional and Deep Anomaly Detection methods. 
\\

\noindent In summary, this work provides the following contributions:

\begin{enumerate}
\item We introduce the robust outlier detection algorithm CoMadOut, whose baseline variant (CMO) derives margin-based decision boundaries by utilizing the noise-resistant eigenpairs of comedian PCA.
\item We proposed additional CoMadOut variants (CMO*), which improve outlier scoring by simultaneously considering measures of out-of-distribution (tailedness) in order to enhance comedian PCA based scoring to arbitrary distributions, e.g. by kurtosis weighted scores of CoMadOut variant CMO+k.
\item We detail, how the proposed CMO* variants can be combined to an ensemble approach for outlier detection (CoMadOut variant CMOEns).
\item We conduct extensive experiments comparing and discussing the performance of our methods against competitors and several real-world datasets.
\end{enumerate}



\noindent The remaining work is structured as follows. In section~\ref{sect:relwork}, we give an overview of related work. Section~\ref{sect:comadOut} introduces our CoMadOut outlier detection method and elaborate several variants of this method. The experiments are presented in section~\ref{sect:exp}, while section~\ref{sect:conclusion} concludes the paper.

\section{Related Work}
\label{sect:relwork}

In this section, we review the related literature, concerning all approaches intersecting with our approach CoMadOut. Thereby we divide those into three categories: (1)~well established traditional outlier detection methods, (2) deep outlier detection methods, and (3)~robust estimation and PCA-based methods, where the latter are optimized towards outlier robustness most similar to our proposed method~CoMadOut.

\subsection{Traditional Methods}

The Local Outlier Factor (LOF)\cite{LOFbreunigKriegel} is a density-based outlier detection method. As such it determines if an object is an outlier based on the $k$NN-neighborhood. Samples which exhibit a significantly lower density in their own local neighborhood compared to the density of other samples and their respective neighborhoods are identified as outliers.

The Isolation Forest (IF)\cite{IsoForestLiu} method is a tree ensemble approach to identify anomalies. The decision trees of that ensemble are initially constructed by randomly selecting a feature and then performing a random split between its minimum and maximum value recursively. IF is based on the assumption that outliers exhibit a lower occurrence in contrast to "normal" samples making them appear close to the root of a tree with fewer splits necessary.

The One Class SVM (OCSVM)\cite{ocsvmCrammer} approach separates all samples from the origin by maximizing the distance from a separating hyperplane to the origin. Therefore, kernels can be utilized to transform the samples into a high dimensional and thus better separable space. Consequently, a binary function computes which regions in the original data space exhibit a high density and are therewith labelled with "+1" while all other samples~(outliers) are labelled with "-1".

\subsection{Deep Outlier Detection Methods}

Two prominent neural network architectures that are commonly used for deep anomaly detection are Autoencoders~(AE) and Variational Autoencoders (VAE)\cite{An2015VariationalAB},  whose low-dimensional latent vector spaces compete with those of PCA\cite{Hinton2006ReducingTD}. The underlying concept for both is fairly similar. An Encoder is trained to embed each training sample, so that it can be projected into a generally lower dimensional latent space and then reconstructed by a Decoder network. The assumption for anomaly detection is now that if a model is trained well on normal data, samples that are abnormal should be hard to reconstruct with the same network and therefore have a high reconstruction error. A threshold is then used to identify these anomalies. Among the different variations of AE, VAEs are special versions of autoencoders. VAEs first encode the input as distributions over the latent space followed by a sampling of samples from the learned distributions. In essence VAEs fit normal distributions on the data and achieve a separation from abnormal samples. Another well established deep anomaly detection algorithm is DeepSVDD (Deep Support Vector Data Description)~\cite{pmlr-v80-ruff18a}. Following the unsupervised AD assumption, that the majority of instances in the training set is expected to be normal, their approach learns the weights of a neural network with the goal to reduce the dimensionality of its inputs and maps these embeddings towards a center~$c$ in the representation space while simultaneously minimizing an enclosing hypersphere around~$c$ by using quadratic loss and weight decay regularization. The abnormality of representations is scored by their distance to~$c$.

\subsection{Robust Estimation and PCA-based Methods}
 


Within this work PCA (Principal Component Analysis) of \citeauthor{JolliffePCA} is denoted as standard PCA. It is a well-known technique to find patterns in high dimensional data by analysing the variances within the data. As stated in previous chapters, its outlier-sensitive mean of the calculated covariance matrix makes it sensitive towards outliers and thus it is not ideal for subsequent tasks like outlier detection. Thus, a more robust, mean-free version is required to perform reliably on outlier datasets.

Comedian PCA\cite{outlcomedianapproach,DBLP:conf/sisap/KazempourH019} follows the same goal as other outlier resistant robust PCA methods and the core idea behind it is intriguingly simple: Instead of computing the eigendecomposition of the covariance matrix, which is highly susceptible to outliers, the computation is performed on a comedian matrix~\cite{Falk1997}. Analogously to the covariance matrix, the comedian matrix represents the median absolute deviation from the median for each dimension and each respective pair of dimensions. Therefore, the coMAD-PCA considers the components with the highest deviation to the median, while standard PCA captures the deviation to the mean. Since the mean of standard PCA is generally more sensitive to outliers~\cite{OutlierSensitivityOfPCA}, coMAD-PCA should be less sensitive. That has been shown by~\citeauthor{outlcomedianapproach} and~\citeauthor{DBLP:conf/sisap/KazempourH019}, who illustrated the robustness of coMAD-PCA towards outliers.


Amongst the methods that also consider robustness is the so called Minimum Covariance Determinant (MCD)\cite{rousseeuw1984least}. The goal of MCD is to find a subset of samples whose covariance matrix has the minimum determinant. Its location is then the subset's average, and its variance is the subset's covariance matrix. \citeauthor{rousseeuw1984least} states in addition that this technique can yield suitable results even when 50$\%$ of the data are contaminated with outliers. 

Since the performance of Empirical Covariance Estimators, like e.g. the performance of the Maximum Likelihood Covariance Estimator (MLE), suffers from distorted eigenvectors when used on datasets with outliers, more robust methods have been developed. They replace the outlier sensitive parts mean and (co)variance by robust alternatives. These alternatives use e.g. the samples of the lowest covariance matrix determinant (MCD) or randomly selected samples (FastMCD)\cite{Rousseeuw98afast}\footnote{\label{fn:fastmcd}cf. Elliptic Envelope (\url{https://scikit-learn.org/stable/modules/generated/sklearn.covariance.EllipticEnvelope.html})} to provide a robust mean and covariance or compute it deterministically (DetMCD)\cite{Hubert10adeterministic} in order to achieve a robust distance measure and thereby a high breakdown value\footnote{\label{fn:breakdownvalue}proportion of outlier samples an estimator can handle before returning a wrong result}.
 
With the goal of becoming robust to outliers, there were several ideas on how to achieve this. At this point the role of coMAD for CoMadOut (cf. section~\ref{sect:comadOut} - step~1) shows parallels to the Stahel–Donoho outlyingness~(SDO) measure\cite{StahelDonohoOutlyingnessMeasure}, which instead uses a weighted mean vector and covariance matrix, and to the comedian approach\cite{outlcomedianapproach}, which computes robust mahalanobis distances based on a positive-semi-definiteness-corrected comedian matrix and weights them by a $\chi^2$-distribution-factor to receive a suitable cut-off value instead. Further methods with parallels to the role of coMAD are estimators like LMS~(Least Median Squares)\cite{rousseeuw1984least} which fits to the minimal median squared distances and the estimator MOMAD~(Median of Means Absolute Deviation)\cite{Depersin2021OnTR}, which considers the SDO as notion of depth (close to our notion of robust inlierness in section~\ref{sect:CoMadOutStep2}) while estimating mean values, which are robust to outliers. \cite{Pena2001MultivariateOD} detects outliers by locally optimizing projections according to kurtosis coefficients between a contaminated and an uncontaminated distribution (latter used as robust estimator for mean and covariance) using mahalanobis distance and a MAD-normed distance to the median as measure for outlyingness.

PCA-MAD\cite{Huang2021ARA} also address the issue of outlier sensitivity and non-robustness of mean-based PCA by weighting outlier scores based on outlier-sensitive standard PCA projections but MAD-weighted z-score distances instead of outlier resistant and kurtosis-weighted coMAD-PCA projection distances or median based margins as the variants of our approach CoMadOut do~(cf. Fig.~\ref{fig:AUROCPCAMAD}).

With this brief overview on the literature of state-of-the-art outlier detection methods we differentiated existing methods from our approach.
Best to our knowledge there are no other approaches like our CoMadOut baseline~(cf. section~\ref{sect:comadOut} CMO Steps 1-5) using inlier region enhancing coMAD-based orthogonal distance medians as robustness improving noise margins in combination with coMAD-based robust inlier regions to identify, score and predict outliers. Furthermore, there is no approach like our CoMadOut variants CMO* optimizing coMAD-PCA based outlier scores by simultaneously weighting according to variance or tailedness~(cf. section~\ref{sect:CoMadOutStep3CMO*} CMO* Steps 1-3).\\

\section{CoMadOut}
\label{sect:comadOut}
\begin{figure}[hbt!]
\includegraphics[width=\textwidth]{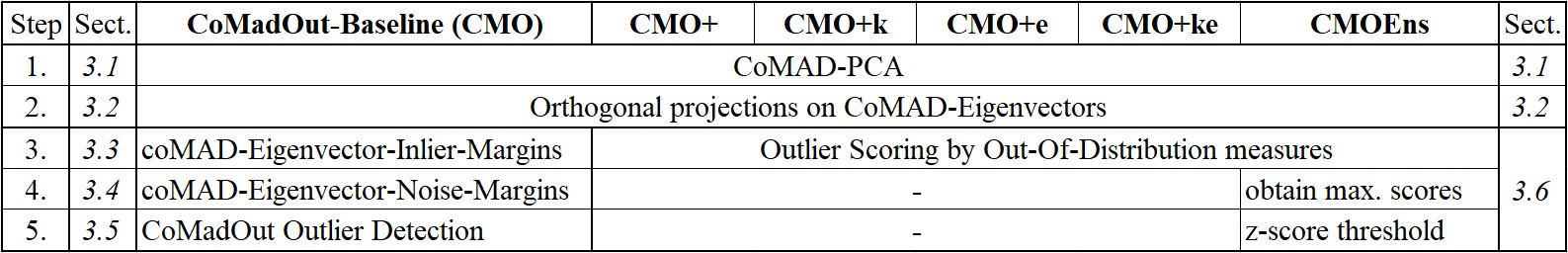}\centering
\caption[]{Overview of CoMadOut (CMO*) variants in bold and their steps.}
\label{fig:Fig14CMOOverview}
\end{figure}

With this section we introduce CoMadOut~(CMO), an unsupervised robust outlier detection algorithm, which follows the assumption that coMAD-based principal components and a robust measure of in-distribution (ID), median~$m$, as well as measures of out-of-distribution (OOD), kurtosis $\kappa$, optimize the alignment of the decision boundary between normal and abnormal instances. Thus we introduce besides the baseline algorithm CMO also its variants CMO*. The following paragraphs provide a step-wise explanation and address similarities and differences between the several CMO variants. An overview is provided by the table in Fig.~\ref{fig:Fig14CMOOverview}.

\begin{figure}[ht!]
\begin{minipage}[b]{.55\linewidth}\centering
\subfloat[coMAD-PCA]{\label{fig:CoMADPCA}\includegraphics[width=\linewidth]{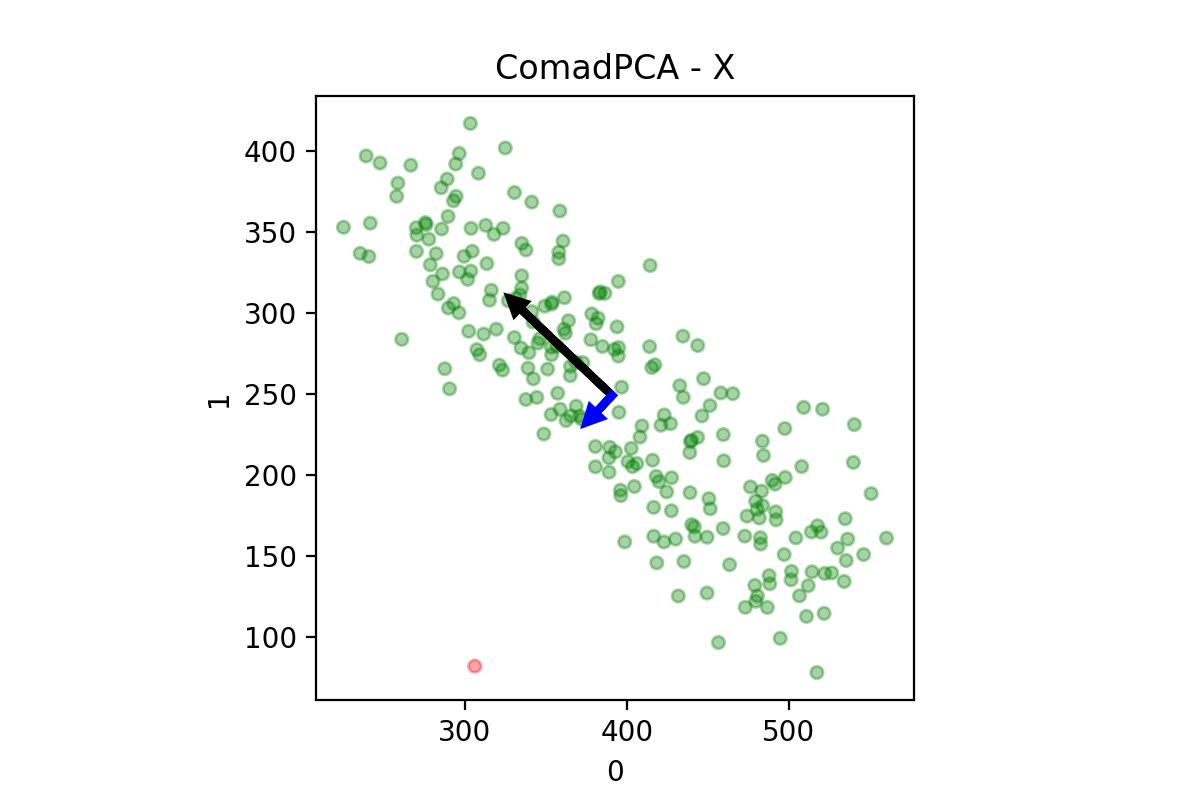}}
\end{minipage}
\hspace{-1.5cm}
\begin{minipage}[b]{.55\linewidth}\centering
\subfloat[standard PCA]{\label{fig:StandardPCA}\includegraphics[width=\linewidth]{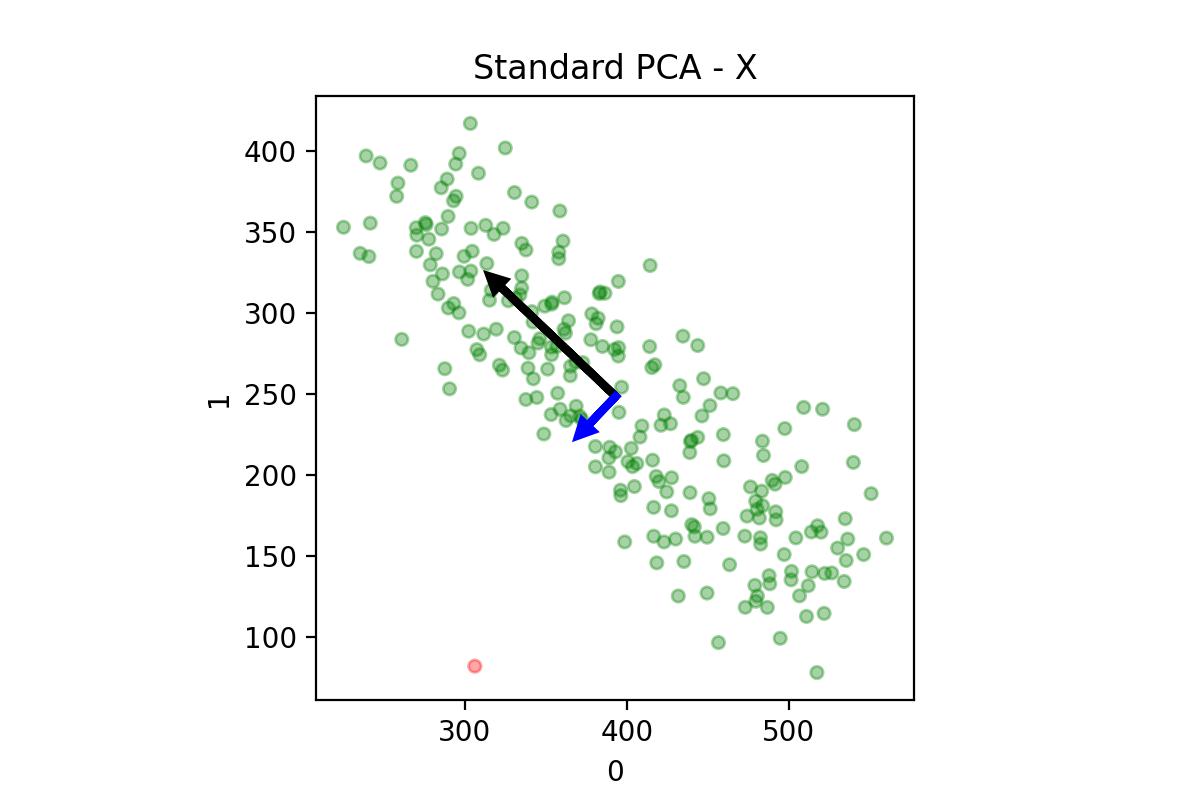}}
\end{minipage}
\caption[]{Illustration of orientation and scale of principal components after applying coMAD-PCA (a) and standard PCA (b) to a synthetic dataset. It can be observed that the outlier (colored in red) has less influence to eigenvectors and eigenvalues of coMAD-PCA on the left than to those of standard PCA on the right.}
\label{fig:CoMadPCAvsStandardPCA}
\end{figure}

\noindent The common goal of all CoMadOut variants is to first of all obtain outlier resistant subspace orientations. This is achieved by (1) using coMAD-PCA\cite{outlcomedianapproach,DBLP:conf/sisap/KazempourH019} with their robust comedian matrix\cite{Falk1997} instead of standard PCA with its outlier sensitive covariance matrix~(cf. Fig.~\ref{fig:CoMadPCAvsStandardPCA}). This allows CoMadOut to utilize the outlier resistant eigenvectors and eigenvalues of coMAD-PCA in order to receive a robust subspace representation~(a coordinate system with axes spanned by the coMAD-PCA eigenvectors) and with that the initial region for inliers for the baseline approach CMO. Since noise is often modeled as a weak form of outliers~\cite{10.5555/3086742}, CoMadOut improves its outlier robustness and its false positive rate also by (2) an additional noise margin~(NM) based on the median of the euclidean orthogonal distances between the samples and their subspace axes $k$ extending the initially calculated inlier region. 
\\

\noindent In particular CoMadOut can be described by the following steps and variants. All variants have step 1 and 2 in common and get variant-specifc starting from step 3, whereas the width of a cell indicates for which algorithm variant in the column header its content is relevant.

\FloatBarrier

\subsection{Step 1 - Computation of CoMAD-PCA Matrix}\label{sect:CoMadOutStep1}

Let $X$ be a centered matrix with $n$ samples $x \in \{x_{1}, x_{2} , ..., x_{n}\}$ having $d$-dimensional feature vectors in $\mathbb{R}^{n\times d}$ and let the comedian\footnote{due to the ambiguous usage of COM for co-median and comedian matrix the term coMAD matrix from\cite{DBLP:conf/sisap/KazempourH019} has been used interchangeably} matrix $COM(X)$ of \cite{Falk1997} be defined by

\begin{equation}
\begin{tabular}{c c}
$\scriptstyle X = 
\begin{pmatrix}
    x_{1,1} & \hdots & x_{1,d} \\
    \vdots & \ddots & \vdots \\
    x_{n,1} & \hdots & x_{n,d} \\
\end{pmatrix},
$
&
$\scriptscriptstyle COM(X) = 
\begin{pmatrix}
    com(A_{1},A_{1}) & \hdots & com(A_{1},A_{d})\\
    \vdots & \ddots & \vdots \\
    com(A_{d},A_{1}) & \hdots & com(A_{d},A_{d})\\
\end{pmatrix}$\\
\end{tabular}
\end{equation}

\noindent with $A_{k}$ being the $k$-th feature of samples $x_{*,k}$ and with co-median
\begin{equation}
com(A_{i},A_{j}) = med((A_{i} - med(A_{i}))(A_{j} - med(A_{j})))
\end{equation}
\noindent so that the comedian matrix $COM(X)$ acts as counterpart for the covariance matrix $\Sigma$ in the original PCA, whereas 
\begin{equation}
A_{i} - med(A_{i})
\end{equation}
\noindent represents the robustness providing subtraction of the median in place of the subtraction of the outlier sensitive mean.
\noindent On the resulting comedian matrix we apply PCA like~\cite{outlcomedianapproach,DBLP:conf/sisap/KazempourH019} in order to utilize its resulting robust eigenpairs to define the selective inlier region for the CMO baseline. Therefore, we consider 
\begin{equation}
\delta(X) = U \Lambda ~U^T
\end{equation}
\noindent with eigenvector matrix $U$ and eigenvalue matrix $\Lambda$ (=$COM(X)$), where those eigenpairs~(consisting out of eigenvector and eigenvalue) with the $k$-largest eigenvalues are used for the definition of the initial inlier region (for CMO baseline) in the next step 2. Due to the known non-positive-semi-definiteness of the comedian matrix a part of the score-relevant eigenvalues is negative. The approach of \citeauthor{outlcomedianapproach} addressed this issue by replacing the original comedian-based eigenvectors and eigenvalues of $\delta$ with iteratively adjusted projections resulting in an approximated but positive-semi-definite surrogate matrix. 
For our CoMadOut approaches we do not want to lose the properties of the original comedian matrix. Therefore, we keep the original non-positive-semi-definite comedian matrix and assume symmetrical gaussian distributed projections so that the sign of the usually small part of negative eigenvalues can be neglected just for the scope of final outlier scoring. Hence, the magnitude of all eigenvalues can get equally included into outlier scoring by taking their absolute value and adjusting the scoring accordingly. 
\begin{equation}
\lambda = \textpipe diag(\Lambda) \textpipe
\end{equation}
This allows us to base our outlier scoring still on the original robust eigenvectors by additionally including a small adapted eigenvalue part. 

Since the ideal choice of $K$ for the number of principal components can vary between datasets, we conducted the experiments of our work (cf. section~\ref{sect:exp}) on different percentages of the total number of $L$ possible subspaces (leading to a different number of $K$ but maximal $L$ principal components) for all datasets and averaged the achieved average performances. 

\subsection{Step 2 - Orthogonal projections on coMAD-Eigenvectors}\label{sect:CoMadOutStep2}
In the scope of CoMadOut the previously calculated eigenpairs are used to calculate the scores of inlierness and outlierness respectively. Therefore, the orthogonal projections of all samples $x$ are calculated for each eigenvector~(=principal component). The eigenpairs define the direction ($k$-th eigenvector $\vec{u_{k}}$ of matrix~$U$) and scale ($k$-th eigenvalue $\lambda_{k}$ of matrix~$\Lambda$) and correspond to a subspace axis $k$. 

\noindent Formally step 2 can be described by the following equations:
$\forall x \in X$ the projections $x'$ can be calculated by
\begin{equation}
x' = ((x^T u_{k}) / (u_{k}^T u_{k})) \cdot u_{k}
\end{equation}
\noindent Since the origin of the subspace is zero centered after coMAD-PCA from Step~1, the euclidean distances of the projected samples $x'$ to the origin can be calculated as
\begin{equation}
||x'||_{k} = \sqrt {\sum_{k=1}^{K}  \left( x_{k}'\right)^2 } 
\end{equation}

\noindent After the first two steps, which have CMO baseline approach and its variants CMO* in common, the next steps are either related to CMO or to its CMO* variants.

\subsection{CMO: Step 3 - Computation of coMAD-Eigenvector-Inlier-Margins}\label{sect:CoMadOutStep3CMO}

The next step for the margin-based CMO baseline approach is to span an initial region for inliers. The eigenpairs define the direction ($k$-th eigenvector $\vec{u_{k}}$ of matrix~$U$) and scale ($k$-th eigenvalue $\lambda_{k}$ of matrix~$\Lambda$) of the corresponding eigenvectors, which correspond to the subspace axes. In order to find out which samples $x$ are part of the inlier region, only samples are considered, which got projected on subspace axis $k$ between its origin $O$ and the absolute length of its axis-corresponding eigenvector $\vec{u_{k}}$. Since the eigenvector originally is an unit vector the actual eigenvector length~$||\vec{u_{k}}||$ -~and with that the inlier range for each subspace axis - is defined by the scale of the related eigenvalue $\lambda_{k}$. That $\lambda_{k}$ spans the range of our inlier region in positive and negative eigenvector direction with $[-\lambda_{k},+\lambda_{k}]$ on each subspace axis $k$.

\noindent Formally, the decision for each sample $x'$, if it lays within the inlier range of all subspace axis eigenvectors $\vec{u_{k}}$, is defined as follows:
\begin{equation}
inl(x')_{k}= 
\begin{cases}
    1,& \text{if}~||x'||_{k} \in [-\lambda_{k},+\lambda_{k}]\\
    0,              & \text{otherwise}
\end{cases}
\end{equation}
\noindent In order to avoid inliers to be considered as outliers when the inlier region defining eigenvalue $\lambda_{k}$=0 a minimum inlier region epsilon $\epsilon$ of 1e-6 is defined. 

\noindent In case a sample $x'$ lays within all inlier ranges the product of all subspace-axis-k-related inlier function results 
\begin{equation}
inl(x') = \lceil \frac{ \sum_{k=1}^K {inl(x')_{k}} }{K} \rceil
\end{equation}
\noindent would be equal to 1 (=inlier), otherwise 0 (=non-inlier, cf. section~\ref{sect:CoMadOutStep5CMO} step 5).

\subsection{CMO: Step 4 - Computation of coMAD-Eigenvector-Noise-Margins}\label{sect:CoMadOutStep4CMO}

\noindent After defining robust inliers in step 3 also noise samples\footnote{\label{fn:noisesamples}in this work samples are considered as noise samples when they are closer to the related subspace axis than its orthogonal distances median $m_{k}$ but not part of the set of initial robust inliers defined by the coMAD-eigenpairs (cf. section~\ref{sect:CoMadOutStep2} step 2)} are considered for adding them to the set of inliers. Therefore, we extend the robust inlier region ranges, which are corresponding to their subspace axis $k$
\begin{equation}
[-\lambda_{k},+\lambda_{k}]
\end{equation}
by the median of its euclidean orthogonal distances representing a robust noise margin bandwidth $m_{k} = med(||x'||_{k})$ leading to new inlier region ranges
\begin{equation}
[-m_{k}-\lambda_{k},+\lambda_{k}+m_{k}].
\end{equation}
Thereby, the restrictive $\lambda$-thresholds receive an extended margin for each principal component $k$, capable to also cover noise samples as inliers by still using the outlier robust distances median instead of the outlier sensitive distances mean.

\subsection{CMO: Step 5 - CoMadOut Outlier Detection}\label{sect:CoMadOutStep5CMO}

\noindent Having the final decision boundaries between inliers and outliers available (cf. step~4), this step describes how the CMO baseline approach is able to \textit{identify}, \textit{score} and \textit{predict} samples $x$ as outliers.

Since the coMAD-PCA enables CoMadOut to create its eigenpairs according to the robust coMAD matrix, outliers are less likely to distort their orientation and scale, which allows to reliably detect inliers in the first phase. Further restrictively selecting noise samples as inliers increases the set of robust inliers and increases thereby the probability that the remaining samples are actual outliers, which as a consequence enables CoMadOut to implicitly identify outliers.

CoMadOut is also capable to provide sample outlier scores $sc_{i}$. For this purpose the absolute euclidean distances $x'_{k}$ of the projected samples $x'_{i}$ are first reduced by the inlier margin $\tau_{k}$ aiming that actual outlying non-negative distance residuals of the principal components remain. Based on own empirical evaluations the best representation for the final outlier score $sc_{i}$ for a sample $x'_{i}$ is the mean of its remaining non-negative distance residuals.

\begin{equation}
||sc_{i}|| = mean(max(0,||x'_{i}||_{k} - \tau_{k}))
\end{equation}

CoMadOut can also provide softmax scores. For that the median from all $\lambda$-scaled scores $sc_{i}$ for each score dimension $k$ is calculated to represent an in-distribution score. The non-negative subtraction of that score median from all sample scores allows to uncover residual score dimensions which are more likely to represent an outlier sample so that the subsequent softmax function returns for these dimensions higher outlier scores. The highest softmax score among them represents the softmax score of the related sample $x'_{i}$. 

\begin{equation}
||sc_{i,softmax}|| = max_{i}(softmax(sc_{i,k}/\sqrt{\lambda_{k}}))
\end{equation}
\newline

\noindent Finally the outlier prediction of the CMO baseline is realized by spanning a hyperrectangle or a $k$-dimensional rectangular cuboid (for $k$=2, 3, ..., $K$ principal components) with the help of the outlier resistant coMAD-eigenpairs according to the related scale ($k$-th eigenvalue~$\lambda_{k}$ of matrix~$\Lambda$) and direction ($k$-th eigenvector $\vec{u_{k}}$ of matrix~$U$) respectively, which allows to separate between inliers and outliers by outlier threshold $\tau$ with $\tau_{k} = \lambda_{k} + m_{k}$
\begin{equation}
outl(x')_{k}=
\begin{cases}
    1,& \text{if}~||x'||_{k} \notin [-\tau_{k},+\tau_{k}]\\
    0,& \text{otherwise}
\end{cases}
\end{equation}
\noindent In case a sample $x'$ exceeds at least one of the $K$ given subspace axis outlier thresholds $\tau_{k}$, the outlier decision function

\begin{equation}
outl(x') = \prod_{k=1}^K {outl(x')_{k}}
\end{equation}

\noindent would be equal to 1 (=outlier) and otherwise~0 (=inlier).\\

\begin{figure}[ht!]
\includegraphics[width=0.65\textwidth]{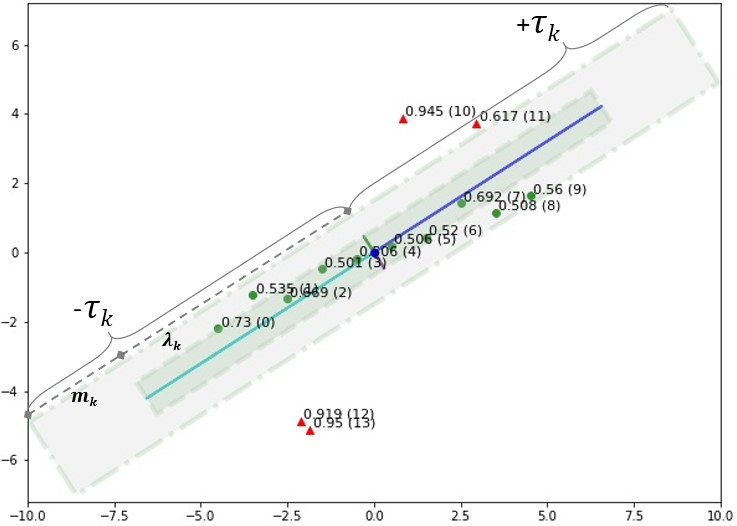}\centering
\caption[]{CoMadOut scoring and labeling of CMO baseline on synthetic data.}
\vspace{-0.5cm}
\label{fig:Fig2CMO}
\end{figure}

For a geometrical intuition of the scoring and labeling of the CoMadOut baseline approach CMO please see~Fig.~\ref{fig:Fig2CMO}. In~Fig.~\ref{fig:Fig2CMO} CMO is visualized on a 2-dimensional test dataset. The first principal component (PC) is depicted by the blue and turquoise lines in positive and negative direction and the second PC by the purple and green line accordingly. They define a hyperrectangle (or a $k$-dimensional rectangular cuboid for $K\ge3$) for the robust inlier region~(green background) and get extended by the noise margin (NM)(gray background), which defines the boundary between inliers (green) and outliers (red).
\vspace{-1.5cm}

\FloatBarrier

\subsection{CMO*: Step 3 - Outlier Scoring considering weighting according to  out-of-distribution measures}\label{sect:CoMadOutStep3CMO*}

Since the outlier prediction based on the strict inlier and outlier regions of the rectangular cuboid of the CoMadOut baseline approach CMO~(cf. Fig.~\ref{fig:Fig2CMO}) is partially too restrictive for a variety of high-dimensional datasets, further variants of CoMadOut CMO* are introduced with this work. Still utilizing the beneficial properties of coMAD-PCA~(cf. sections~\ref{sect:CoMadOutStep1} and~\ref{sect:CoMadOutStep2} with steps 1 and~2) like outlier resistant eigenpairs or the location of the median as measure of in-distribution, the predictions of variants CMO* are pure outlier scores softening the strict inlier and outlier regions of the CMO baseline.  

The related but different approach PCA-MAD of~\citeauthor{Huang2021ARA} addressed this issue by weighting outlier scores based on outlier-sensitive standard PCA projections (instead of outlier resistant coMAD-PCA projections) and based on principal component based MAD- and median-scaled z-scores. 

Our approach CMO+ defines outlier scores based on the sum of absolute distances of coMAD-PCA projections to their zero centered origin~$O$ without weighting the outlier scores.

\begin{figure}[ht!]
\centering
\includegraphics[width=0.4\textwidth]{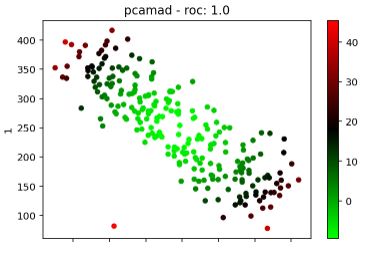}
\caption[]{Outlier score boundaries of competitor PCA-MAD\cite{Huang2021ARA}.}
\label{fig:AUROCPCAMAD}
\end{figure}

\begin{figure}[ht!]
\includegraphics[width=\textwidth]{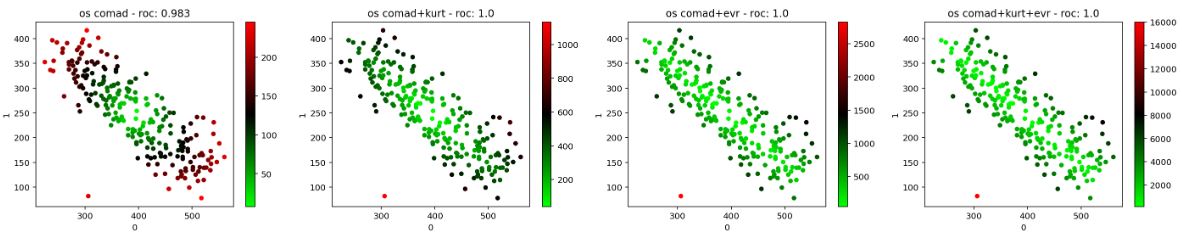}
\caption[]{Outlier score boundaries of CoMadOut variants CMO+, CMO+k, CMO+e, CMO+ke on a synthetic dataset (from left to right).}
\label{fig:AUROCCMOvariants}
\end{figure}

Comparing CMO+~(cf. Fig.~\ref{fig:AUROCCMOvariants}) and PCA-MAD\cite{Huang2021ARA}~(cf. Fig.~\ref{fig:AUROCPCAMAD}) one recognizes strong and weak points on both sides. Although the coMAD-based distribution center of CMO+ seems to be well centered and its boundaries are not as geometrically restrictive as those of the CMO baseline it only considers samples quite close to the distribution median as normal. Depending on the domain-specific anomaly semantics of a dataset such outlier scoring could be too restrictive as well. PCA-MAD of \citeauthor{Huang2021ARA} has a broader range of normal non-outlier scores around its MAD-scaled mean but struggles like CMO+ to align its outlier scores close to the distribution boundaries of the longitudinally shaped dataset. Scores of border points close to the mean are still representing non-outliers whereas already points in the middle between boundary and mean start to get scored as outliers.

In order to address the issue of neglected distribution boundaries, further CoMadOut variants introduce OOD measures to weight its CMO* outlier scores ($sc_{i} = x_{i}'$) accordingly. CMO+k introduces kurtosis based weighting of outlier scores with 
\begin{equation}
sc_{i_{CMO+k}} = \kappa \cdot sc_{i} 
\end{equation}
with kurtosis $\kappa$ as a measure of dispersion between the two $\mu$ and $\sigma$ \cite{moors1986meaning}, and provides with that a notion of distribution tailedness.
\\\\
Formally kurtosis $\kappa$ is according to Pearson defined as:\\
 
\begin{equation}
\quad \kappa = \mathbb{E} \left[ \left( \frac{X - \mu}{\sigma} \right) ^{4} \right] = \frac{\mu^{4}}{\sigma^{4}}    
\end{equation}
where $\mu$ represents the distribution mean and $\sigma$ is the standard deviation.
\\

Considering Fig.~\ref{fig:AUROCCMOvariants}, the kurtosis based weighting of coMAD-PCA based outlier scores $sc_{i}$, introduced with CoMadOut variants CMO+k and CMO+ke, contributes to a distribution boundary focused outlier scoring and thus providing a suitable outlier detection algorithm for related datasets and accordingly demanding domains.

Since \citeauthor{Huang2021ARA} demonstrated with $\lambda$-weighted distance outlier scores as part of PCA-MAD a comparable positive effect for boundary adapted outlier scoring, its performance is also investigated with the coMAD-based CoMadOut variants CMO+e and CMO+ke whereas the latter includes also kurtosis based weighting.

Formally the outlier scoring of variants CMO* is defined as follows:
 
\begin{equation}
sc_{i_{CMO+e}} = \frac{sc_{i}}{\lambda}  
\end{equation}

\begin{equation}
sc_{i_{CMO+ke}} = \frac{\kappa \cdot sc_{i}}{\lambda} 
\end{equation}

Considering the CMO* scoring in Fig.~\ref{fig:AUROCCMOvariants}, the applied $\lambda$-weighting as introduced in \cite{Shyu03anovel} with $\lambda$ as weighting factor, leads to a permanent notion of normality along the direction of the main principal component so that even boundary points in that direction are considered as normal. Comparing the outlier scoring for boundary points, $\lambda$-weighted scoring shows due to the principal component related variance less focus to the outlier scores of boundary points in comparison to the pure kurtosis weighted outlier scores.

Investigating also in the combined performance of all CoMadOut variants CMO* this work also introduces its ensemble approach CMOEns which standardizes the outlier scores of CMO* methods by its z-score

\begin{equation}
zsc_{i_{*}} = \frac{sc_{i_{*}} - \mu_{*}}{\sigma_{*}} 
\end{equation}

\noindent and selects for each sample the largest z-score among the outlier z-scores of the CMO* approaches. 

\begin{equation}
zsc_{i_{CMOEns}} = max(zsc_{i_{CMO+}},zsc_{i_{CMO+k}},zsc_{i_{CMO+e}},zsc_{i_{CMO+ke}}) 
\end{equation}

\noindent For the ensemble variant CMOEns a sample is considered as an outlier when z-score $zsc_{i_{CMOEns}}$ for a sample exceeds the outlier threshold of $z_{thresh}=1$: 

\begin{equation}
outl(zsc_{i_{CMOEns}})=
\begin{cases}
    1,& \text{if}~zsc_{i_{CMOEns}} \notin [-z_{thresh},+z_{thresh}]\\
    0,& \text{otherwise}
\end{cases}
\end{equation}

With this section~\ref{sect:comadOut} we introduced the CoMadOut variants CMO and CMO* whose performances are investigated in scope of extensive experiments in the next section~\ref{sect:exp} comparing them to the performance of related and state of the art methods for unsupervised anomaly detection.
\FloatBarrier

\section{Experiments}
\label{sect:exp}

In order to demonstrate the competitiveness of the CoMadOut variants CMO and CMO*, several experiments have been conducted to measure its capability to detect outliers as well as its runtime behavior.

\subsection{Setup}
\label{sect:expsetup}

For our experiments we used a paperspace GPU instance with Linux Ubuntu 20.04.3 LTS~(Focal Fossa), Intel(R) Core(TM) i7-6500U CPU @ 2.50GHz QuadCore and 45GB RAM memory with usage of 48GB GPUs and parallel processing. Moreover, we wrote our experimental code in Python using Jupyter Notebook. 

\subsubsection{Datasets} First we want to show how datasets could look like on which our CoMadOut approach works well and on which it starts to fail. For this purpose we created a synthetic toy dataset shown in Fig.~\ref{fig:CoMadPCAvsStandardPCA} with 1 extreme outlier and 235 instances as part of the main distribution of normal instances. We
investigated and discussed the algorithm robustness in terms of coMAD-PCA based outlier resistance (cf. details provided with Fig.~\ref{fig:AUROCPCAMAD} and Fig.~\ref{fig:AUROCCMOvariants} in Section~\ref{sect:CoMadOutStep3CMO*}). 
In order to further extend the experiments with CoMadOut we also benchmarked the variants of our method on \textit{real world datasets} like the well-known 20newsgroups dataset\footnote{\label{fn:news20}\url{http://qwone.com/~jason/20Newsgroups/}}representing a large high-dimensional dataset, Boston Housing Prices\footnote{\label{fn:housingdataset}\url{https://archive.ics.uci.edu/ml/machine-learning-databases/housing/}} dataset, the PARVUS Wine dataset\footnote{\label{fn:winedataset}\url{http://odds.cs.stonybrook.edu/wine-dataset/}} and many other datasets~(cf. Tab.~\ref{tabledatasets}) from the ODDS benchmark website\footnote{\label{fn:odds}\url{http://odds.cs.stonybrook.edu}} providing more samples and features than the initial synthetic dataset. 
The Boston Housing Price dataset has been extended by a ground truth label column "outlier" replacing the original regression value target column "MEDV". Therefore and to receive a preferably linear dataset in case of outlier removal, we considered samples laying 2 Inter Quartile Ranges (IQRs) out of the 1st and 3rd quartile as outliers and flagged them with~"1" whereas the normal data got flagged with~"0". Moreover, the Wine dataset got specifically tailored for the task of outlier detection. In this version from ODDS\cref{fn:winedataset} samples with label~"class~1"~(cultivator~1) are reduced to~9 samples and flagged as outliers with label~"1". Samples with labels~"class~2" or~"class~3" are considered as normal samples and flagged as inlier with label~"0". 
Since CMO variants calculate their comedian matrices including all given samples, the 20newsgroup\cref{fn:news20} dataset had to be downsampled to 10\% from (n: 11.905, p: 768) to (n: 1.302, p:768) to not exceed the hardware and runtime restrictions but it still represents an high-dimensional real-world dataset.

\begin{table*}[hbt!]
\centering
\caption[]{Overview of datasets used in our experiments.}
\label{tabledatasets}
\resizebox{\textwidth}{!}{%
\tiny
\begin{tabular}{|l|r|r|r|}
\hline
dataset & \multicolumn{1}{c|}{samples} & \multicolumn{1}{c|}{features} & \multicolumn{1}{c|}{outliers} \\ \hline
20news         & 1300          & 768         & 65 (5.0\%)                    \\ \hline
arrhytmia         & 451          & 274         & 65 (14.4\%)                    \\ \hline
cardio         & 1830          & 21         & 176 (9.6\%)                    \\ \hline
annthyroid         & 7199          & 6         & 534 (7.4\%)                    \\ \hline
breastw         & 682          & 9         & 239 (35.0\%)                    \\ \hline
letter         & 1599          & 32         & 100 (6.3\%)                    \\ \hline
thyroid         & 3771          & 6         & 93 (2.5\%)                    \\ \hline
mammography         & 11182          & 6         & 260 (2.3\%)                    \\ \hline
pima         & 767          & 8         & 267 (34.8\%)                    \\ \hline
musk         & 3061          & 166         & 96 (3.1\%)                    \\ \hline
optdigits         & 5215          & 64         & 150 (2.9\%)                    \\ \hline
pendigits         & 6869          & 16         & 156 (2.3\%)                    \\ \hline
mnist         & 7602          & 100         & 700 (9.2\%)                    \\ \hline
shuttle         & 49096          & 9         & 3510 (7.1\%)                    \\ \hline
satellite         & 6434          & 36         & 2036 (31.6\%)                    \\ \hline
satimage-2         & 5802          & 36         & 71 (1.2\%)                    \\ \hline
wine         & 128          & 13         & 9 (7.0\%)                    \\ \hline
vowels         & 1455          & 12         & 50 (3.4\%)                    \\ \hline
glass         & 213          & 9         & 9 (4.2\%)                    \\ \hline
wbc         & 377          & 30         & 21 (5.6\%)                    \\ \hline
boston         & 506          & 14         & 179 (35.4\%)                    \\ \hline
\end{tabular}
}
\end{table*}

\subsubsection{Compared Methods}
\label{sect:methods}

With this paper we also want to benchmark our approach against a wide range of state-of-the-art outlier detection algorithms in order to introduce our method as a further competitive and robust alternative to already well established outlier detection algorithms and to emphasize the performance of CoMadOut compared to some of its most similar approaches like PCA-MAD\cite{Huang2021ARA}\footnote{\label{fn:githublohrerapcamad}\url{https://github.com/lohrera/pcamad}}, Elliptic Envelope\cref{fn:fastmcd} and Minimum Covariance Determinant~(MCD)\cite{Rousseeuw98afast, rousseeuw1984least}. For the comparison to non- and semi-robust algorithms we also involved several variants of standard PCA methods (PCA from \citeauthor{Shyu03anovel} and its mean-sensitive covariance matrix in combination with several scoring methods: raw scores - PCA(r), eigenvalue-based scores - PCA(e), scores based on our robust noise margin - PCA(NM), our kurtosis weighted scores - PCA(k) and their combination PCA(ke)). The latter four allow us to directly compare between the impact of comedian and standard PCA for the scoring, once based on robust eigenpairs of CMO(*) approaches and once based on mean-sensitive eigenpairs of PCA(*) approaches.
In addition to that also traditional outlier detection methods like LOF\cite{LOFbreunigKriegel}, KNN\cite{10.1007/s10618-015-0444-8}, IsolationForest(IF)\cite{IsoForestLiu}, HBOS\cite{Goldstein2012HistogrambasedOS} and OCSVM\cite{ocsvmCrammer}, as well as modern Deep Outlier Detection methods like AutoEncoder(AE)\cite{Hinton2006ReducingTD}, Variational AE(VAE)\cite{An2015VariationalAB} and DeepSVDD\cite{pmlr-v80-ruff18a} had been involved in our benchmark experiments. As far there existed already reliable implementations of the algorithms in python frameworks like scikit-learn\footnote{\label{fn:scikitlearn}\url{https://scikit-learn.org}}, PyOD\footnote{\label{fn:pyod}\url{https://pyod.readthedocs.io}} or others\cref{fn:githublohrerapcamad} then those had been used. In total, 5 random-states and seed values had been set from 0 to 4 for all experiments, allowing approaches with random effects converge to a stable result. The hyperparameters had been set as follows for the related datasets (cf. Tab.~\ref{tabledatasets}). CoMadOut~(softmax\_scoring=False, center\_by='median'), PCA-based (n\_components=0.25, 0.999), AE (hidden\_neurons=[4,3,2,2,3,4], batch\_size=4, dropout\_rate=0.0, epochs=10, l2\_regularizer=0.01 for all), VAE (encoder\_neurons=[4, 3, 2], decoder\_neurons=[2, 3, 4], batch\_size=4, epochs=10, dropout\_rate=0.0, l2\_regularizer=0.001) and DeepSVDD (use\_ae=True, hidden\_neurons=[64, 32, 4], batch\_size=32, epochs=10, dropout\_rate=0.0, l2\_regularizer=0.1). For all other parameters default values had been used for all experiments. For DeepSVDD we recognized that given default parameters are not sufficient to demonstrate the potential of the approach on tabular data, which could be further improved in future work with dataset specific hyperparameter optimization. So far available, the benchmark results of PCA-MAD\cite{Huang2021ARA} had been used directly from their paper, otherwise from the non-official implementation\cref{fn:githublohrerapcamad}. Evaluation metrics described in the subsequent paragraph had been measured with its scikit-learn\cref{fn:scikitlearn} implementations.

\subsubsection{Evaluation Metrics}
\label{sect:evalmetrics}

Outlier detection can be a semi- or unsupervised task conducting binary decisions whether a given sample is normal~(=inlier or noise) or abnormal~(=outlier). Since the algorithms are mostly threshold-based and the ratio between normal and abnormal data is more likely to be skewed, we use the evaluation metrics Average Precision (AP), Area Under the Receiver Operating Characteristic~(AUROC) curve, Area Under Precision Recall Curve (AUPRC) and Precision@n (P@n) with $n$ as the number of total samples. 

Especially Precision and Recall allow in presence of highly skewed data a more accurate view on an algorithms performance.\cite{precrecallroc} 

Depending on the relevance for a given domain problem, different evaluation metrics should be considered. In case an anomaly detection method should avoid wrong alerts (false positives) metrics like AUROC should be considered, whereas domains aiming to avoid missed alert (false negatives) should involve AUPRC in their performance analysis. Due to the usually highly imbalanced datasets, metrics like accuracy neglecting an explicit analysis of false predictions are not involved in our evaluation.

\FloatBarrier
\subsection{Results}
\label{sect:results}
After explaining the experimental setup we introduce our experiment results. The performance of the CoMadOut variants has been compared with that of~26 other outlier detection methods which have been benchmarked with the datasets and evaluation metrics described in section~\ref{sect:expsetup}. Since the performance of the algorithms based on standard PCA and comedian PCA is dependent on the number of chosen principal components the benchmark can be run on different percentages of the total number of principal components. In order to demonstrate the differences between extreme percentages of e.g. 0.25 (25\%) and 1.0 (100\%), both benchmark results are reported for each performance metric starting with the results on all principal components (100\%).

The metrics in these tables (cf. Appendix~\ref{sect:secA1}) represent the performance for each combination between algorithm and dataset as well as the average performances~("AVG") of the particular algorithms as well as on a specific dataset. Their row "WIN" indicates how many times an algorithm has achieved the best performance from all benchmarked algorithms. Providing a notion of how much better an algorithm performs compared to all other algorithms the row "ARK" contains Average Rank, which each algorithm achieved on the tested datasets (lower ARK means better). The last row "RK" contains dense absolute ranks starting with 1 as rank for the best algorithm. The cells highlight the best result per algorithm in bold. Results are referenced in brackets related to its paragraph e.g. in paragraph "Results AP \& AUROC" the reference (4/8) means (AP-RK100\%: 4/AP-RK25\%: 8).

Average ranks are the basis for the conducted pair-wise Wilcoxon test with Friedman-Nemenyi post-hoc tests investigating if an algorithm or algorithm group performs significantly better than others or shows no significant performance differences. Inspired by ADBench~\cite{NEURIPS2022_cf93972b}, we calculate a critical difference $cd$ based on the Wilcoxon approach with a significance level of $\alpha=0.05$, which allows us to illustrate which algorithms perform equally well or better (cf. e.g. the critical difference diagram in the following Fig.~\ref{fig:Fig6ap0999cddiagram}, where bold line ranges indicate no significant difference between algorithms of this range for the given experimental setting and difference value). The critical difference diagram splits the algorithm list into two halfs, the best performing algorithm on the upper left and the comparably worst performing on the upper right. The average rank "ARK" is shown in the diagram right next to an algorithm providing a notion of how much better or worse an algorithm is compared to another one.
\FloatBarrier

\subsubsection{Results AP \& AUROC}
\begin{figure}[ht!]
\includegraphics[width=\textwidth]{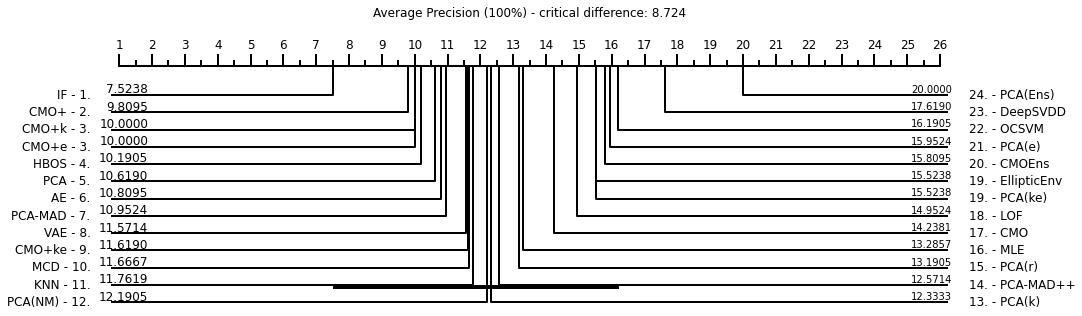}\centering
\caption[]{Average Ranks of Average Precision (AP) performances based on 100\% of all possible principal components. Algorithms within bold black bars show no significant better performance. CMO* are ours, 1. is the best.}
\label{fig:Fig6ap0999cddiagram}
\end{figure}

\begin{figure}[ht!]
\includegraphics[width=\textwidth]{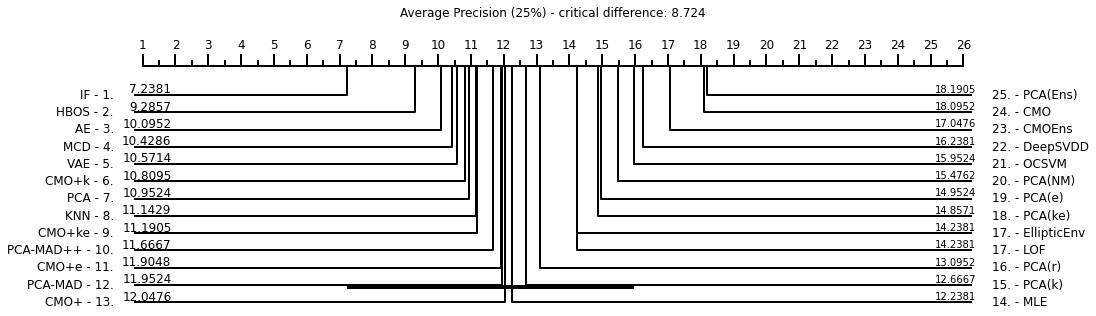}\centering
\caption[]{Average Ranks of Average Precision (AP) performances based on the top 25\% of all possible principal components. Algorithms within bold black bars show no significant better performance. CMO* are ours, 1. is the best.}
\label{fig:Fig10ap025cddiagram}
\end{figure}

Considering the results related to the evaluation metrics AP~(cf. Tab.~\ref{tab:Fig6ap0999} and Tab.~\ref{tab:Fig10ap025}) CoMadOut variants perform best ("win") on 5 datasets out of 21 (for 100\% and 25\% respectively), e.g.~CMO+ (2/13) shows on the second best AP rank behind Isolation Forest (1/1) and together with CMO+k (3/6) and CMO+e (3/11) the best performance among all compared PCA-based methods for 100\% of all principal components (PCs). Also noticeable is the performance of the standard PCA variants with CoMadOut scoring ("PCA(NM,k,ke,Ens)") on top of covariance matrices achieving wins on 5 and 3 datasets for 100\% and 25\% PCs. The related critical difference (cd) diagrams in Fig.~\ref{fig:Fig6ap0999cddiagram} and Fig.~\ref{fig:Fig10ap025cddiagram} confirm these findings, but according to the Nemenyi post-hoc test the performances between the best algorithm IF with ranks (1/1) and e.g. OCSVM with ranks (22/21) at the end of the critical difference range cannot be considered as significantly better ($\alpha>0.05$). In addition to that, the cd diagrams indicate an insignificant performance loss of many CMO* variants after PC reduction from 100\% to 25\%, except the kurtosis-weighted variants (CMO+k, CMO+ke), which are comparably stable in terms of less average rank loss and demonstrate with that also its suitability for robust outlier detection after dimensionality reduction.

\begin{figure}[ht!]
\includegraphics[width=\textwidth]{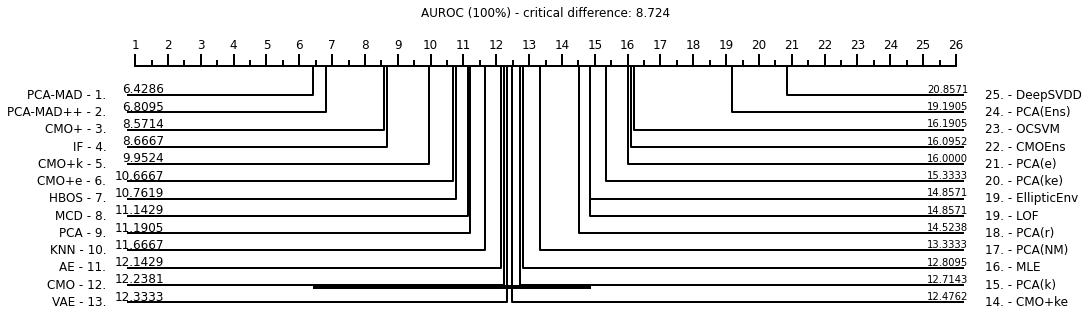}\centering
\caption[]{Average Ranks of AUROC performances based on 100\% of all possible principal components. Algorithms within bold black bars show no significant better performance. CMO* are ours, 1. is the best.}
\label{fig:Fig4roc0999cddiagram}
\end{figure}

\begin{figure}[ht!]
\includegraphics[width=\textwidth]{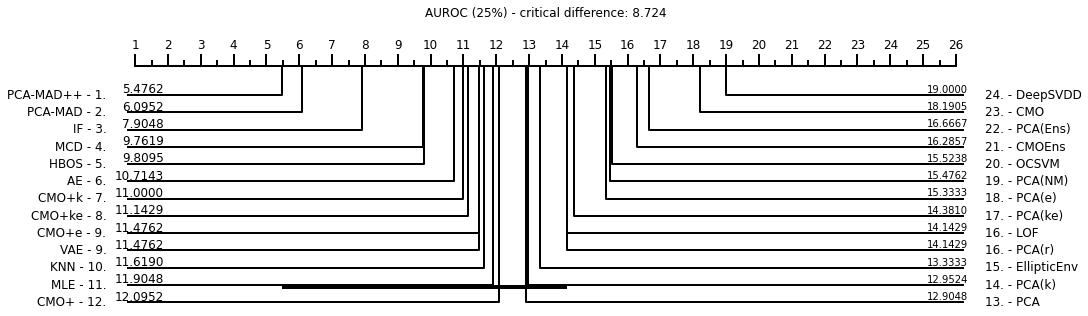}\centering
\caption[]{Average Ranks of AUROC performances based on the top 25\% of all possible principal components. Algorithms within bold black bars show no significant better performance. CMO* are ours, 1. is the best.}
\label{fig:Fig9roc025cddiagram}
\end{figure}

For AUROC the CoMadOut variants achieve in total 6 best dataset performances on 100\% PCs and 5 on 25\%. The variants CMO+ (3/12), CMO+k (5/7) and CMO+e (6/9) show the best CMO performances and in case of the first two comparable many total wins as the PCA-MAD (1/2) but are on par with Isolation Forest (4/3) on 100\% PCs~(cf. Tab.~\ref{tab:Fig4roc0999}). On only the top 25\% of all principal components the performance of CoMadOut is comparable with those from the upper midfield~(Tab.~\ref{tab:Fig9roc025}). Considering the cd diagrams (cf. Fig.~\ref{fig:Fig4roc0999cddiagram} and Fig.~\ref{fig:Fig9roc025cddiagram}) PCA-MAD variants outperform all algorithms in each AUROC setting, whereas on 100\% PCs the CMO* variants and IF are right behind and on 25\% PCs methods as IF, MCD and HBOS. PCA(*) variants also demonstrate by total wins (5/3) that the combination of outlier sensitive covariance matrices and CoMadOut scoring can be beneficial for a selection of datasets, which also includes with 20newsgroup one of the large-scale real-world datasets. 

\FloatBarrier

\subsubsection{Results AUPRC \& Precision@n}

\begin{figure}[ht!]
\includegraphics[width=\textwidth]{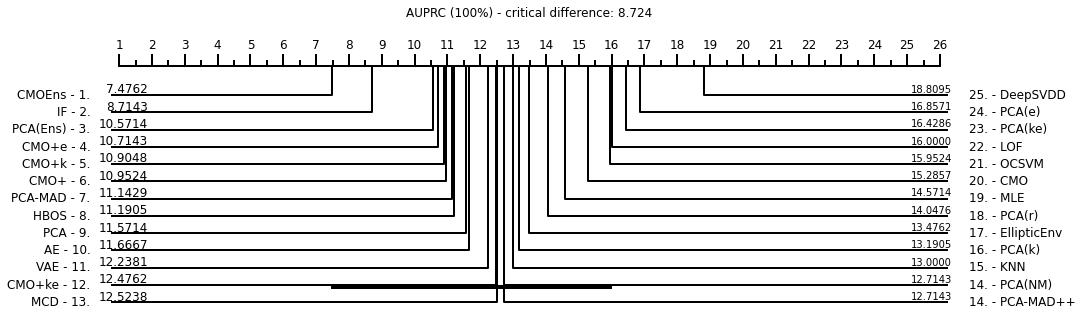}\centering
\caption[]{Average Ranks of AUPRC performances based on 100\% of all possible principal components. Algorithms within bold black bars show no significant better performance. CMO* are ours, 1. is the best.}
\label{fig:Fig10recall0999cddiagram}
\end{figure}

\begin{figure}[ht!]
\includegraphics[width=\textwidth]{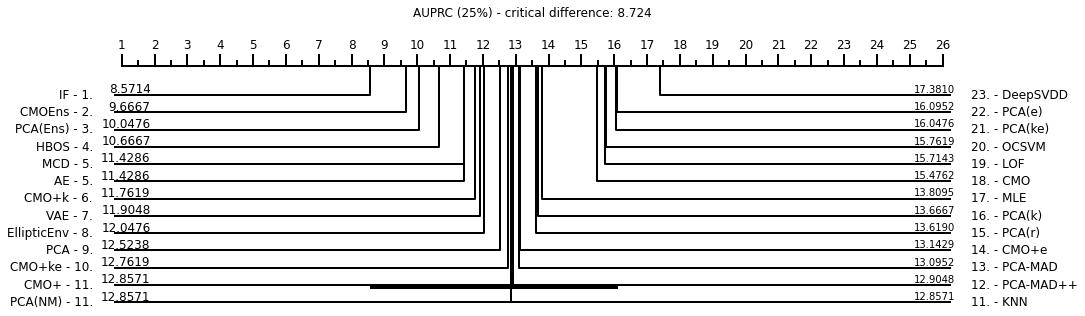}\centering
\caption[]{Average Ranks of AUPRC performances based on the top 25\% of all possible principal components. Algorithms within bold black bars show no significant better performance. CMO* are ours, 1. is the best.}
\label{fig:Fig11recall025cddiagram}
\end{figure}

Comparing the AUPRC performances of the CoMadOut variants, one observes that the CoMadOut variant CMOEns shows the best and second-best performance (1/2) of all algorithms, being on par with Isolation Forest (2/1), and 13 total wins could be achieved by all CMO variants~(cf. Tab.~\ref{tab:Fig10recall0999} and Tab.~\ref{tab:Fig11recall025}). Furthermore, the scoring variants of CoMadOut could also enhance the scoring for standard PCA methods leading to a total of 7 and 3 wins and PCA(Ens) as third-best (3/3) of all algorithms. Also the performance of CMO+, CMO+k and CMO+e are ranked in the top six on 100\% PCs. For comparison of CMO* approaches with all other algorithms the related cd diagrams are shown in Fig.~\ref{fig:Fig10recall0999cddiagram} and Fig.~\ref{fig:Fig11recall025cddiagram}.

\begin{figure}[ht!]
\includegraphics[width=\textwidth]{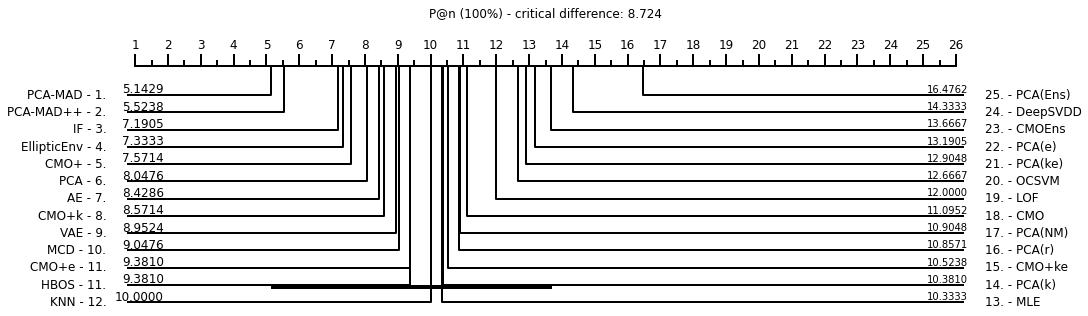}\centering
\caption[]{Average Ranks of Precision@n (P@n) performances based on 100\% of all possible principal components. Algorithms within bold black bars show no significant better performance. CMO* are ours, 1. is the best.}
\label{fig:Fig7precn0999cddiagram}
\end{figure}

\begin{figure}[ht!]
\includegraphics[width=\textwidth]{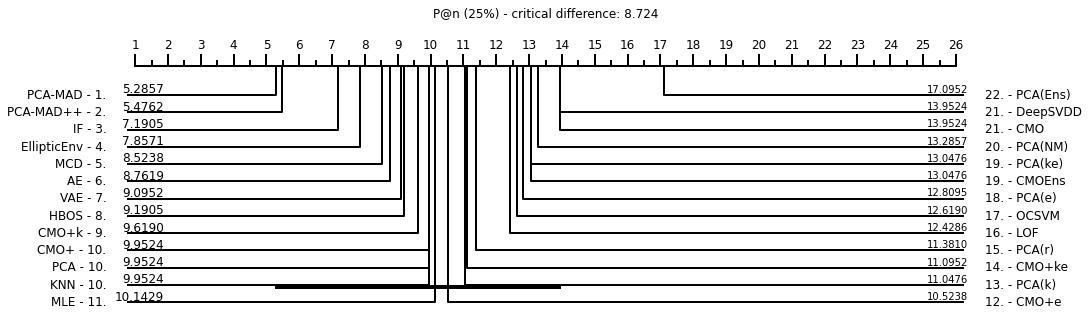}\centering
\caption[]{Average Ranks of Precision@n (P@n) performances based on the top 25\% of all possible principal components. Algorithms within bold black bars show no significant better performance. CMO* are ours, 1. is the best.}
\label{fig:Fig12precn025cddiagram}
\end{figure}

Considering the performances for Precision@n (P@n) in Tab.~\ref{tab:Fig7precn0999} and Tab.~\ref{tab:Fig12precn025} the CoMadOut variants achieved a total of 11 best dataset performances on 100\% and 25\%. Most remarkable is the standard PCA performance with kurtosis weighting of CoMadOut, PCA(k), with a total of 6 best dataset performances. Furthermore interesting are the best dataset performances of methods CMO+k (8/9) and CMO+ (5/10) on 20newsgroup dataset demonstrating its potential for large high-dimensional datasets. The best overall P@n performances achieved PCA-MAD (1/1), PCA-MAD++ (2/2) and Isolation Forest (3/3). Despite these findings the results for the P@N experiments of CMOEns and PCA(Ens) require with zero values further validation.
For a detailed comparison of CMO* approaches with all other algorithms the related cd diagrams are shown in Fig.~\ref{fig:Fig7precn0999cddiagram} and Fig.~\ref{fig:Fig12precn025cddiagram}.



\subsubsection{Runtime}

\begin{figure}[ht!]
\includegraphics[width=\textwidth]{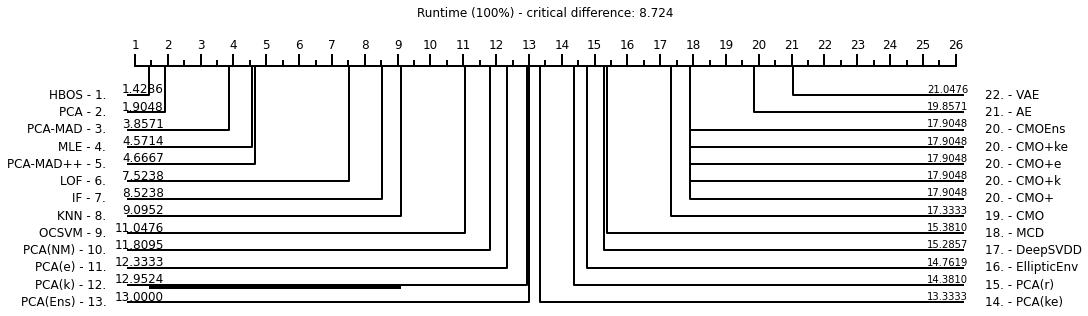}\centering
\caption[]{Average Ranks of Runtime performances based on 100\% of all possible principal components. Algorithms within bold black bars show no significant better performance. CMO* are ours, 1. is the best.}
\label{fig:Fig8runtime0999cddiagram}
\end{figure}

\begin{figure}[ht!]
\includegraphics[width=\textwidth]{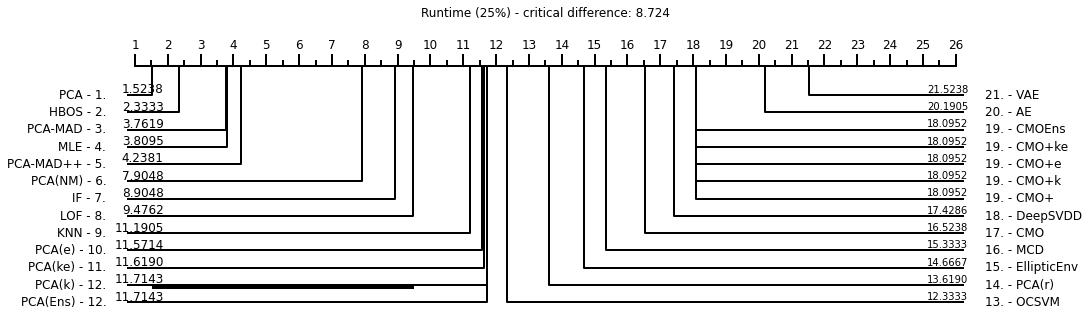}\centering
\caption[]{Average Ranks of Runtime performances based on the top 25\% of all possible principal components. Algorithms within bold black bars show no significant better performance. CMO* are ours, 1. is the best.}
\label{Fig:Fig13runtime025cddiagram}
\end{figure}

Considering the runtime performances of the CoMadOut variants in Tab.~\ref{tab:Fig8runtime0999} and Tab.~\ref{tab:Fig13runtime025} one observes that they tend to show rather lower midfield performances despite parallelization. Comparing the runs of the CMO baseline version with newer CMO* variants one recognizes that the choice of variant depends besides its performances on the previously evaluated metrics also on the amount of principal components whether the baseline algorithm or CoMadOuts newer variants are more reasonable. Independent of the amount of principal components the algorithms HBOS (1/2), PCA (2/1) and PCA-MAD++ (3/3) show the fastest performances. For a detailed comparison of CMO* approaches with all other algorithms the related cd diagrams are shown in Fig.~\ref{fig:Fig8runtime0999cddiagram} and Fig.~\ref{Fig:Fig13runtime025cddiagram}.


\FloatBarrier

\subsubsection{Memory and Runtime Dependencies}

\begin{figure}[ht!]
\begin{minipage}[b]{.49\linewidth}\centering
\subfloat[]{\includegraphics[width=\linewidth]{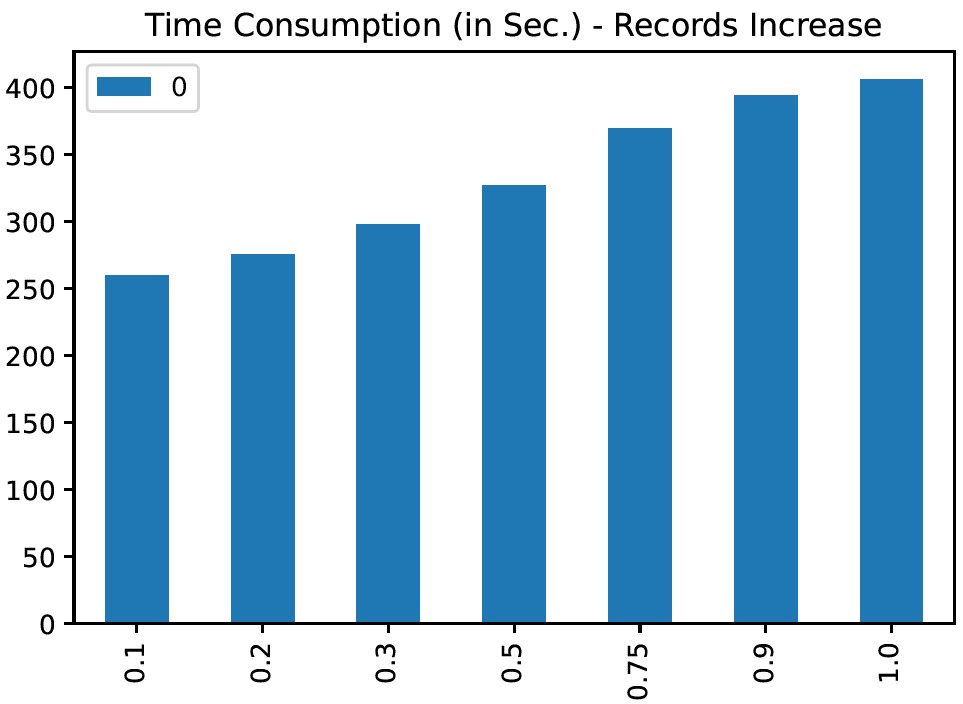}}
\end{minipage}
\begin{minipage}[b]{.49\linewidth}\centering
\subfloat[]{\includegraphics[width=\linewidth]{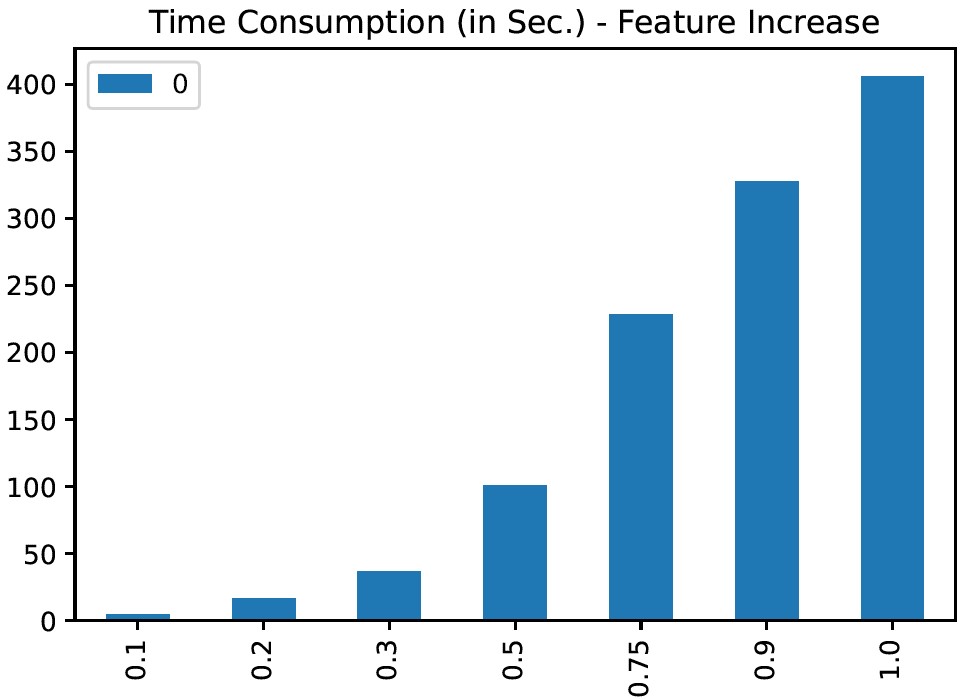}}
\end{minipage}
\caption[]{CMO* time consumption dependent on records and features.}
\label{fig:tc_20news}
\end{figure}

\begin{figure}[ht!]
\vspace{-0.5cm}
\begin{minipage}[b]{.49\linewidth}\centering
\subfloat[]{\includegraphics[width=\linewidth]{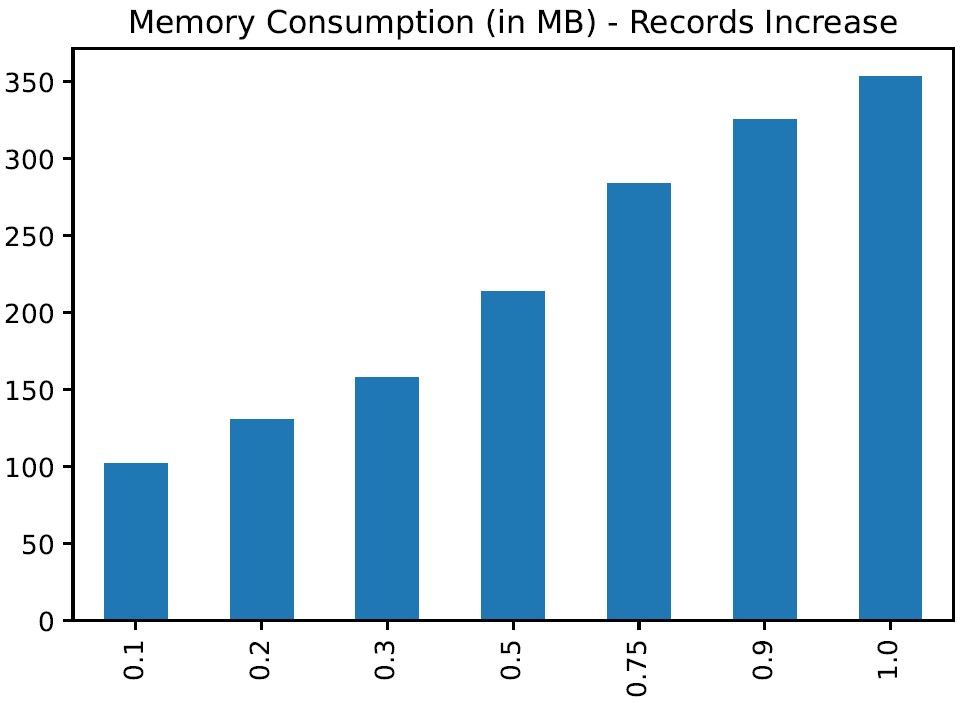}}
\end{minipage}
\begin{minipage}[b]{.49\linewidth}\centering
\subfloat[]{\includegraphics[width=\linewidth]{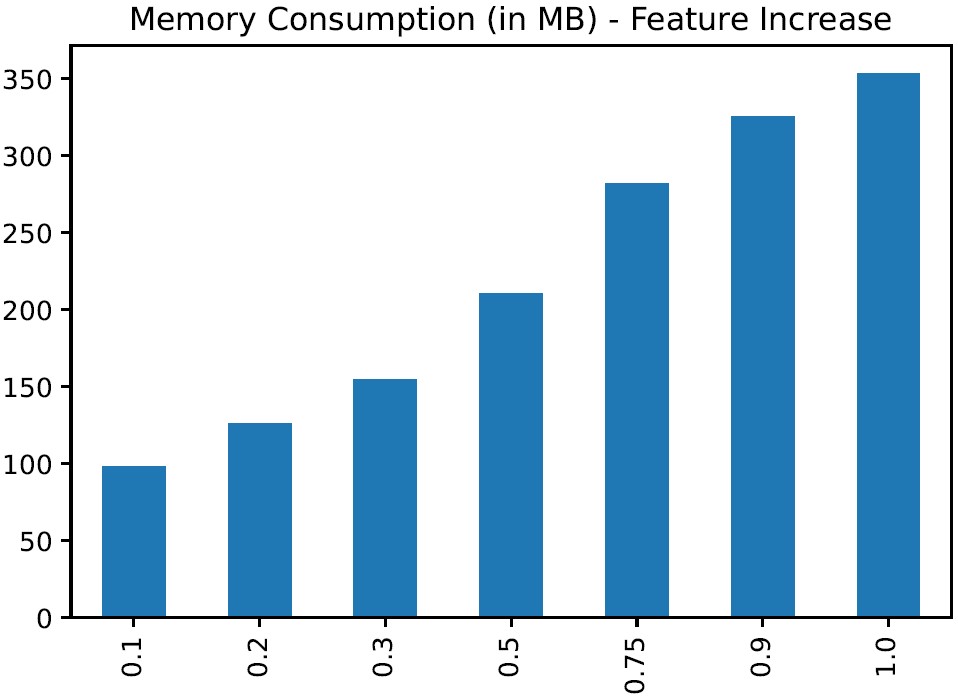}}
\end{minipage}
\caption[]{CMO* memory consumption dependent on records and 
features.}
\label{fig:mc_20news}
\end{figure}

In order to address the dependencies of CMO* variants related to its memory and runtime consumption behavior we made the following investigations. We used with 20newspaper the largest among all compared datasets (11097~samples, 768~features, 222.8MB~raw size) and increased the fraction of samples or features respectively from 0.1 up to 1.0 in each of our tests. In Fig.~\ref{fig:tc_20news} one can observe that in (a) the time consumption is only slightly but constantly increasing with $O(log$ $n)$ by an increasing amount of records, while in (b) the increase of features shows a non-linearly increasing runtime of $O(n^2)$. For the second experiment investigating in memory dependency, one can observe in Fig.~\ref{fig:mc_20news} a) and b) that there is a nearly linear increase of $O(n)$ in memory consumption for both records and feature increase.

\FloatBarrier

\subsubsection{Result Discussion}

The evidence for the contribution of CoMadOut specific \textit{scoring} of inlier- and outlier-regions as part of the CMO approaches is investigated by excluding the outlier-resistant comedian matrix of coMAD-PCA and replacing it with the outlier-sensitive covariance matrix of standard PCA, both done with variants PCA(NM), PCA(r), PCA(k), PCA(e), PCA(ke) and PCA(Ens). Comparing standard PCA based outlier scoring with CMO-based outlier scoring one observes for the latter that PCA(Ens) shows better AUPRC performances (3/3) than PCA (9/9) on all principal components (100\%) but also on the heavily reduced top 25\% principal components in terms of (average) ranks. However, the significance tests in Fig.~\ref{fig:Fig6ap0999cddiagram} and Fig.~\ref{fig:Fig9roc025cddiagram} as well as the comparable average performances show that only the combination of both, outlier-robust dimensionality reduction and CoMadOut scoring methods, allow permanently better performance scores (always with at least one CMO variant ahead) than standard PCA based methods without median usage. 

The evidence for the contribution of outlier scoring based on our nearly unmodified comedian matrices is shown by the CoMadOut variant CMO+ which shows together with the kurtosis-weighted variant CMO+k competitive ranks compared to the majority of all other outlier detection algorithms in terms of AP, AUROC, AUPRC and P@n.
The evidence for the contribution of the distribution tailedness based outlier score weighting is shown when comparing related CoMadOut variant CMO+k with the most related variance- and z-score-weighted approaches of PCA-MAD. In terms of average precision (AP) CMO+k outperforms the related PCA-MAD approaches and shows a competitive overall performance also in terms of AUPRC. However, despite considerable performances of CMO+ and CMO+e, both cannot compete with CMO+k since they neglect to include the kurtosis-based weighting according to the datasets tailedness.

The evidence for robustness is especially shown when comparing CoMadOut variants with outlier-robust methods like PCA-MAD, MCD, etc. on the one side and outlier-sensitive methods like PCA, MLE, etc. on the other side on datasets with a high percentage of outliers and noise (arrhytmia, breastw, pima, satellite, boston). On the majority of these datasets CoMadOut shows compared to the outlier-sensitive methods and partially even also to robust outlier detection methods a better performance in terms of AP, AUROC and AUPRC. Furthermore, the difference is directly visible when comparing the test dataset plots of the principal components of coMAD-PCA and standard PCA next to each other where those of standard PCA are sensitive towards the shown outlier~(cf. Fig.~\ref{fig:CoMadPCAvsStandardPCA} and result tables for details). 

Beside the strengths of our approaches, they also have limitations. On datasets with clear borders between normal and abnormal most CMO variants cannot benefit of its weight-based scoring between normal and abnormal. According to EDA, CMO shows better performances when there are less extreme overlaps within normal and abnormal distributions on the principal components. In the experiments CMO variants showed the best performance on datasets arrhytmia, annthyroid, breastw, thyroid, pendigits, musk, satimage-2 and boston. In addition to that, a restricted variation among the tailednesses of the distributions leads to less effectiveness of variants CMO+k, CMO+e and CMO+ke, whereas reduced noise instance do not allow comedian PCA to show its strength for outlier resistant eigenvectors. A further weakness which has to be paid for fast execution times are the numerical calculations on the whole dataset instead of batch-wise computation leading to high memory consumption and with that to restrictions for hardware and application domains. In addition to that, also the following property identified by \citeauthor{outlcomedianapproach} also hold for our approach due to the comparable comedian matrix basis. In this scope he states the non-positive-semi-definiteness of the comedian matrix. Furthermore, he found breakdown values between outlier ratios $\alpha \in [10,45]$ and identified false detection rates of 0.1\% for symmetrically distributed outliers and 0.7\% for asymmetrically distributed outliers, which need to be also investigated for our approach in future work. Further potential for optimization is the loss of performance of CMO variants which comes along with the reduction of principal components. In this case, CMO+k shows with stable ranks the most robust performances among all other CMO variants.

\subsection{Results based on Parameter-Tuning}
\label{sect:resultsParamTuning}

Although it is not common for an unsupervised setting like ours since one may not have ground truth labels available, we utilize the ground truth labels to address potential performance improvements due to model- and hyperparameter tuning. Therefore and to provide an alternative for commonly used default parameters, we conduct a dataset specific optimization by grid search for all algorithms as far as they offer a sensitive tuning parameter. In this scope we search with seed 0 and the following parameters, value ranges and step sizes listed in Table~\ref{tab:PTranges}. 

\begin{table}[]
\caption{Parameters, ranges and stepsizes used for parameter tuning.}
\begin{tabular}{llll}
\textbf{algorithm or group}       & \textbf{parameter} & \textbf{range}                           & \textbf{stepsize} \\ \hline
PCA-based (CMO*, PCA*)      & n\_components      & {[}0.25, 1.0{]}                          & 0.25              \\
Nearest neighbor based (LOF, KNN) & n\_neighbors       & {[}5, 100{]}                             & 5                 \\
Tree-based (IF)                   & n\_estimators      & {[}10, 100{]}                            & 10                \\
HBOS                              & n\_bins            & {[}10, 100{]}                            & 10                \\
OCSVM                             & nu                 & {(}0, 1{)}                               & 0.1               \\
MCD, EllipticEnv                  & support\_fraction  & {(}0, 1{]}                               & 0.1               \\
AE, VAE, DeepSVDD                 & batch\_size        & $2^{n}$ with n $\in$ {[}2,8{]} & 1                
\end{tabular}
\label{tab:PTranges}
\end{table}

\begin{figure}[ht!]
\includegraphics[width=\textwidth]{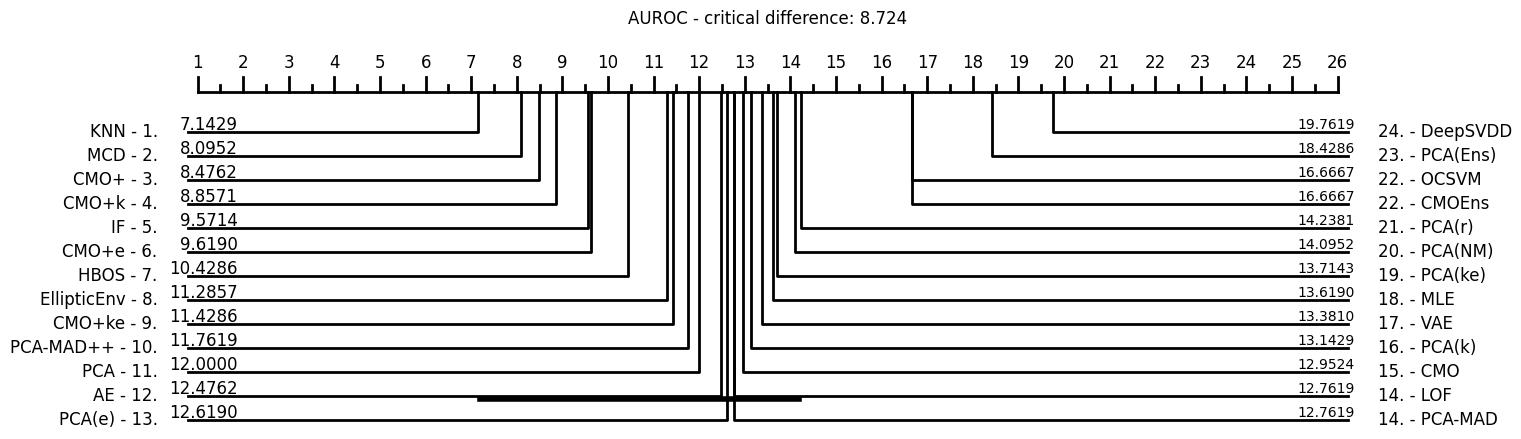}\centering
\caption[]{Average Ranks of AUROC performances based on parameter tuning. Algorithms within bold black bars show no significant better performance. CMO* are ours, 1. is the best.}
\label{Fig:FigPT_auroc_cddiagram}
\end{figure}

\begin{figure}[ht!]
\includegraphics[width=\textwidth]{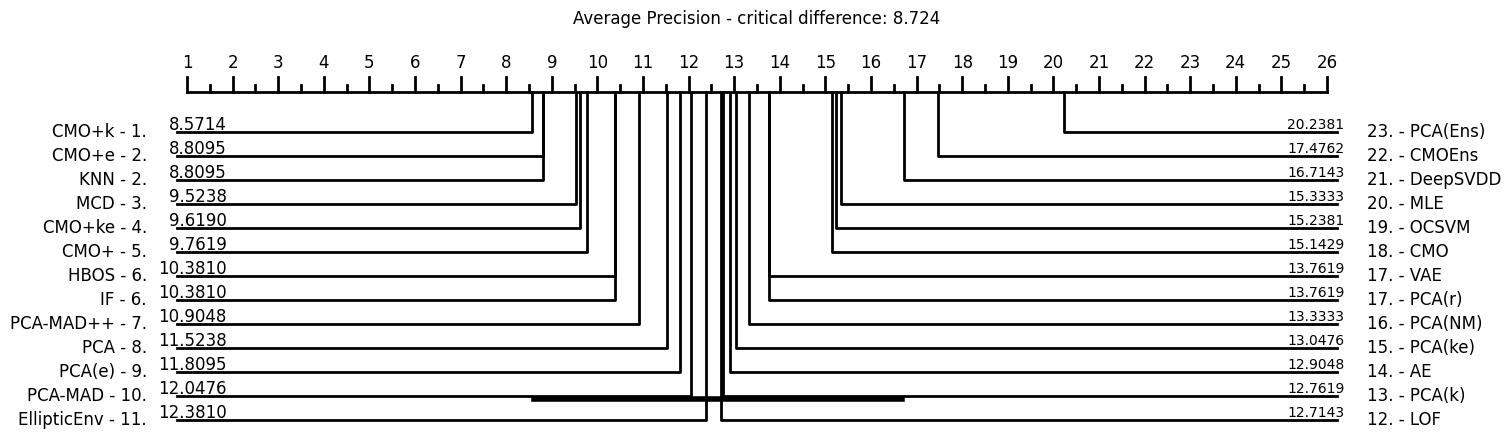}\centering
\caption[]{Average Ranks of Average Precision performances based on parameter tuning. Algorithms within bold black bars show no significant better performance. CMO* are ours, 1. is the best.}
\label{Fig:FigPT_ap_cddiagram}
\end{figure}

\begin{figure}[ht!]
\includegraphics[width=\textwidth]{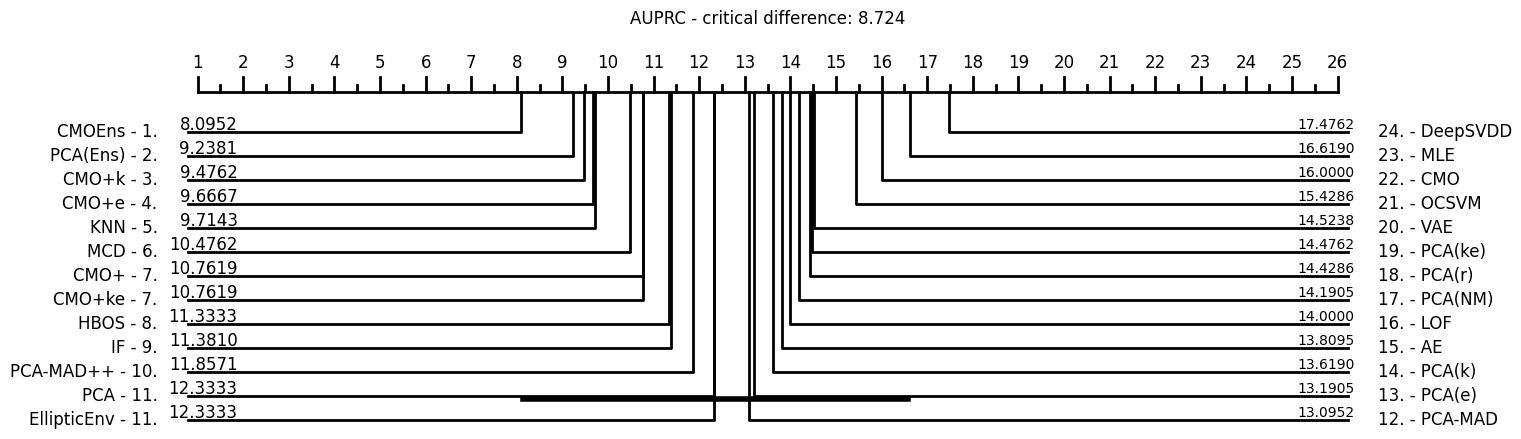}\centering
\caption[]{Average Ranks of AUPRC performances based on parameter tuning. Algorithms within bold black bars show no significant better performance. CMO* are ours, 1. is the best.}
\label{Fig:FigPT_auprc_cddiagram}
\end{figure}

\begin{figure}[ht!]
\includegraphics[width=\textwidth]{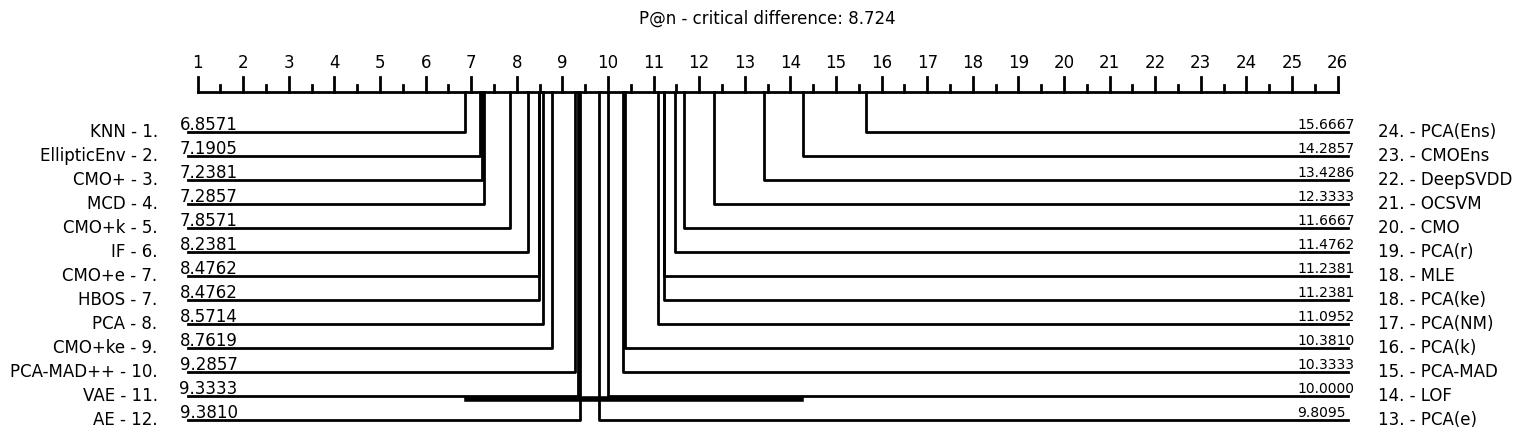}\centering
\caption[]{Average Ranks of P@n performances based on parameter tuning. Algorithms within bold black bars show no significant better performance. CMO* are ours, 1. is the best.}
\label{Fig:FigPT_precn_cddiagram}
\end{figure}

\begin{figure}[ht!]
\includegraphics[width=\textwidth]{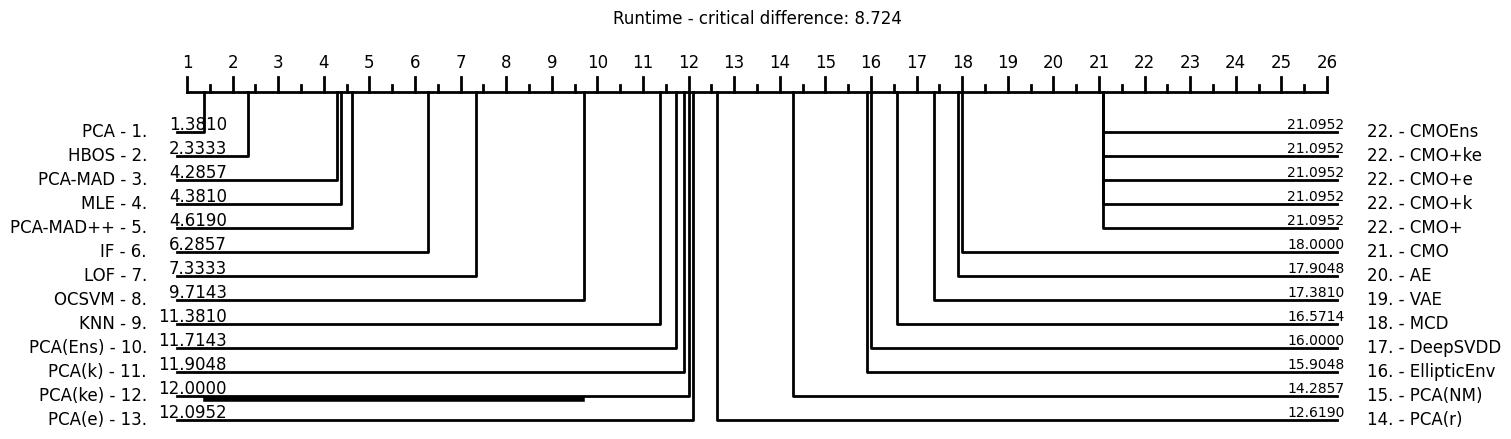}\centering
\caption[]{Average Ranks of runtime performances based on parameter tuning. Algorithms within bold black bars show no significant better performance. CMO* are ours, 1. is the best.}
\label{Fig:FigPT_runtime_cddiagram}
\end{figure}

We optimize the algorithms according to Average Precision (AP) and follow with that metric the approach of~\cite{NEURIPS2021_23c89427}. Accordingly, the best dataset specific algorithm parameters are provided within Table~\ref{tab:FigPTparams}. These by AP selected parameters are the basis for the conducted tests which report the dataset specific algorithm performances by AUROC (cf. Table~\ref{tab:FigPTauroc}), by AP (cf. Table~\ref{tab:FigPTap}), by AUPRC (cf. Table~\ref{tab:FigPTauprc}), by P@n (cf. Table~\ref{tab:FigPTprecn}) and by Runtime (cf. Table~\ref{tab:FigPTruntime}). As already done for the principal component ratio dependent performance analysis we also summarize the algorithm performances by their achieved average rank over all datasets and for each performance metric by critical difference plots, which can be found for AUROC (cf. Figure~\ref{Fig:FigPT_auroc_cddiagram}), for AP (cf. Figure~\ref{Fig:FigPT_ap_cddiagram}), for AUPRC (cf. Figure~\ref{Fig:FigPT_auprc_cddiagram}), for P@n (cf. Figure~\ref{Fig:FigPT_precn_cddiagram}) and for Runtime (cf. Figure~\ref{Fig:FigPT_runtime_cddiagram}).

Compared to the previous analysis KNN shows superior performance in terms of AUROC, AP, AUPRC and P@n. Moreover our CMO+k demonstrates stable on par performances being four times among the top 5 algorithms. Also our CMO+ performs comparably stable being not worse than rank 7. As already demonstrated in earlier analysis, one weakpoint of our CMO* algorithms are worse runtimes which remain as expected also after parameter tuning. In addition to that, our ensemble variant CMOEns still shows good performance for AUPRC. Besides KNN also the algorithm MCD performs comparably well achieving ranks not worse than rank 6 for AUROC, AP, AUPRC and P@n. 

\section{Conclusion}
\label{sect:conclusion}

In this work we proposed CoMadOut, a robust outlier detection algorithm using comedian PCA in order to apply PCA without sensitivity to outliers providing an initially selective inlier region besides an extending inlier noise margin by measures of in-distribution (ID) and measures of out-of-distribution~(OOD). They care for distribution based alignment of the outlier scores of each principal component, and with that, for an appropriate alignment of the decision boundary between normal and abnormal instances. Based on the assumption that samples are only considered as outliers when they are not part of our enhanced inlier region or when they exceed the distribution tailedness weighted outlier score boundary, we demonstrated an on par performance to various kinds of (robust) state-of-the-art outlier detection methods. The consideration of model selection by meta-learners~\cite{NEURIPS2021_23c89427}, of estimating the ideal number of principal components, investigating the performance loss of achieving positive-semi-definiteness while preserving the properties of the original comedian matrix and runtime optimization for CoMadOut in a future work could make our approach even more applicable to high-dimensional datasets and thereby also beneficial to other application domains with the requirement of robust outlier detection.


\newpage
\section*{Author Contributions}
\begin{itemize}
    \item Conceptualization: ALO (Lead), DKA (Supporting), PKR (Supporting)
    \item Methodology: ALO (Lead), DKA (Supporting)
    \item Formal Analysis and Investigation: ALO (Lead)
    \item Implementation: ALO (Lead)
    \item Writing – original draft: ALO (Lead), DKA (Supporting), MAH (Supporting), PKR (Supporting)
    \item Writing – review \& editing: ALO (Lead), DKA (Supporting), MAH (Supporting), PKR (Supporting)
    \item Funding acquisition: PKR (Lead)
\end{itemize}


\section*{Funding}

This work has been funded by the German Federal Ministry of Education and Research (BMBF) under Grant~No.~01IS18036A. The authors of this work take full responsibility for its content.
\section*{Declarations}

\begin{itemize}
\item Employment, Financial interests, Non-Financial interests - Not applicable
\item Conflict of interest - No conflicts or competing interests to declare
\item Ethics approval - Not applicable 
\item Consent to participate/publication - Not applicable (all agree)
\item Employment, Financial interests, Non-Financial interests - Not applicable
\item Availability of data and materials - Not applicable (publicly available)
\item Code availability - planned at \url{https://github.com/lohrera/CoMadOut}
\end{itemize}

\FloatBarrier


\begin{appendices}

\section{Experiment Result Tables}

\label{sect:secA1}


\renewcommand{\arraystretch}{2}
\addtolength{\tabcolsep}{-0.4em}

\begin{sidewaystable}
\caption{Average Precision (AP) performance of 6 CoMadOut (CMO*) variants on the left compared to 20 other anomaly detection algorithms on the right on 21 real world datasets. The results are based on 100\% of all possible principal components.}
\scalebox{0.405}{
\centering
\label{tab:Fig6ap0999}
\begin{tabular}{lrrrrrrrrrrrrrrrrrrrrrrrrrrr}
\toprule 
 & CMO & CMO+ & CMO+k & CMO+e & CMO+ke & CMOEns & \shortstack{PCA-\\[1.5ex]MAD++} & \shortstack{PCA-\\[1.5ex]MAD} & HBOS & IF & PCA & PCA(NM) & PCA(r) & PCA(k) & PCA(e) & PCA(ke) & PCA(Ens) & LOF & KNN & OCSVM & MCD & EllipticEnv & MLE & AE & VAE & DeepSVDD & AVG \\[1.5ex]
\midrule 
20news & 0.044724 & 0.045308 & 0.045306 & 0.046634 & 0.046287 & 0.050000 & 0.045707 & 0.045711 & 0.047568 & 0.045942 & 0.046790 & 0.046427 & 0.047336 & 0.046875 & \bfseries 0.053103 & 0.052243 & 0.048562 & 0.049102 & 0.047434 & 0.045893 & 0.047301 & 0.050000 & 0.045824 & 0.046800 & 0.046733 & 0.048054 & 0.047372 \\[1.5ex]
arrhytmia & \bfseries 0.583627 & 0.562570 & 0.401388 & 0.431523 & 0.436417 & 0.311826 & 0.456989 & 0.489558 & 0.419830 & 0.471999 & 0.395087 & 0.440065 & 0.343874 & 0.342964 & 0.207777 & 0.219795 & 0.202972 & 0.414867 & 0.472525 & 0.136475 & 0.232062 & 0.158957 & 0.285995 & 0.397865 & 0.397781 & 0.169328 & 0.360928 \\[1.5ex]
cardio & 0.251333 & 0.377497 & 0.266436 & 0.397465 & 0.393873 & 0.201126 & \bfseries 0.677366 & 0.567422 & 0.466654 & 0.563401 & 0.612072 & 0.339740 & 0.574698 & 0.343690 & 0.218981 & 0.218981 & 0.293313 & 0.154308 & 0.285894 & 0.533209 & 0.337562 & 0.355416 & 0.463834 & 0.626168 & 0.610493 & 0.113230 & 0.394006 \\[1.5ex]
annthyroid & 0.077931 & 0.625916 & \bfseries 0.662649 & 0.635331 & 0.642549 & 0.270832 & 0.184498 & 0.153584 & 0.183403 & 0.315029 & 0.191271 & 0.074167 & 0.075268 & 0.056714 & 0.077380 & 0.070032 & 0.070823 & 0.205271 & 0.219291 & 0.105847 & 0.504815 & 0.091504 & 0.157324 & 0.231436 & 0.191529 & 0.112233 & 0.237947 \\[1.5ex]
breastw & 0.910604 & 0.987017 & 0.961325 & 0.966530 & 0.863462 & 0.777642 & 0.910820 & 0.957041 & 0.959415 & 0.972006 & 0.957770 & 0.973123 & \bfseries 0.990672 & 0.975286 & 0.935219 & 0.885334 & 0.718703 & 0.271975 & 0.929932 & 0.940377 & 0.968966 & 0.968966 & 0.939455 & 0.847350 & 0.898759 & 0.833327 & 0.896195 \\
\bottomrule 
\end{tabular}

}
\end{sidewaystable}


\renewcommand{\arraystretch}{2}
\addtolength{\tabcolsep}{-0.1em}

\begin{sidewaystable}
\caption{Average Precision (AP) performance of 6 CoMadOut (CMO*) variants on the left compared to 20 other anomaly detection algorithms on the right on 21 real world datasets. The results are based on top 25\% of all possible principal components.}
\scalebox{0.42}{
\centering
\label{tab:Fig10ap025}
\begin{tabular}{lrrrrrrrrrrrrrrrrrrrrrrrrrrr}
\toprule 
 & CMO & CMO+ & CMO+k & CMO+e & CMO+ke & CMOEns & \shortstack{PCA-\\[1.5ex]MAD++} & \shortstack{PCA-\\[1.5ex]MAD} & HBOS & IF & PCA & PCA(NM) & PCA(r) & PCA(k) & PCA(e) & PCA(ke) & PCA(Ens) & LOF & KNN & OCSVM & MCD & EllipticEnv & MLE & AE & VAE & DeepSVDD & AVG \\[1.5ex]
\midrule 
20news & 0.046804 & 0.046544 & 0.046481 & 0.046693 & 0.046697 & 0.050000 & 0.047184 & 0.045714 & 0.047568 & 0.045942 & 0.046818 & \bfseries 0.053226 & 0.048010 & 0.046641 & 0.046462 & 0.046371 & 0.048965 & 0.049102 & 0.047434 & 0.045893 & 0.045775 & 0.050000 & 0.045824 & 0.046800 & 0.046733 & 0.048054 & 0.047374 \\[1.5ex]
arrhytmia & 0.418278 & 0.498537 & \bfseries 0.507573 & 0.473604 & 0.477107 & 0.259604 & 0.437436 & 0.488931 & 0.419830 & 0.471999 & 0.392624 & 0.389153 & 0.261098 & 0.260480 & 0.263084 & 0.250035 & 0.201683 & 0.414867 & 0.472525 & 0.136475 & 0.239083 & 0.158957 & 0.285995 & 0.397865 & 0.397781 & 0.169328 & 0.351690 \\[1.5ex]
cardio & 0.189576 & 0.343230 & 0.380866 & 0.205625 & 0.200570 & 0.149745 & 0.690076 & 0.659719 & 0.466654 & 0.563401 & 0.590827 & 0.255334 & \bfseries 0.703615 & 0.302730 & 0.535627 & 0.263059 & 0.397577 & 0.154308 & 0.285894 & 0.533209 & 0.385621 & 0.344084 & 0.463834 & 0.626168 & 0.610493 & 0.113230 & 0.400580 \\[1.5ex]
annthyroid & 0.091705 & 0.446882 & 0.440181 & 0.437656 & 0.429818 & 0.246400 & 0.083937 & 0.100821 & 0.183403 & 0.315029 & 0.272873 & 0.074167 & 0.052466 & 0.043089 & 0.057144 & 0.043093 & 0.072642 & 0.205271 & 0.219291 & 0.105847 & \bfseries 0.504815 & 0.091504 & 0.157324 & 0.231436 & 0.191529 & 0.112233 & 0.200406 \\[1.5ex]
breastw & 0.678824 & 0.891453 & 0.869548 & 0.843416 & 0.793343 & 0.674565 & 0.876534 & 0.934049 & 0.959415 & 0.972006 & 0.708735 & 0.982058 & \bfseries 0.992965 & 0.992210 & 0.932525 & 0.914502 & 0.704954 & 0.271975 & 0.929932 & 0.940377 & 0.968966 & 0.968966 & 0.939455 & 0.847350 & 0.898759 & 0.833327 & 0.858470 \\
\bottomrule 
\end{tabular}

}
\end{sidewaystable}

\renewcommand{\arraystretch}{2}

\begin{sidewaystable}
\caption{AUROC performance of 6 CoMadOut (CMO*) variants on the left side compared to 20 other anomaly detection algorithms on the right side on 21 real world datasets. The results are based on 100\% of all possible principal components.}
\scalebox{0.42}{
\centering
\label{tab:Fig4roc0999}
\begin{tabular}{lrrrrrrrrrrrrrrrrrrrrrrrrrrr}
\toprule 
 & CMO & CMO+ & CMO+k & CMO+e & CMO+ke & CMOEns & \shortstack{PCA-\\[1.5ex]MAD++} & \shortstack{PCA-\\[1.5ex]MAD} & HBOS & IF & PCA & PCA(NM) & PCA(r) & PCA(k) & PCA(e) & PCA(ke) & PCA(Ens) & LOF & KNN & OCSVM & MCD & EllipticEnv & MLE & AE & VAE & DeepSVDD & AVG \\[1.5ex]
\midrule 
20news & 0.451249 & 0.445444 & 0.446216 & 0.464179 & 0.459844 & 0.496761 & 0.448359 & 0.465799 & 0.463108 & 0.457231 & 0.471417 & 0.458050 & 0.475428 & 0.472949 & \bfseries 0.532370 & 0.525842 & 0.479352 & 0.494002 & 0.472295 & 0.468315 & 0.453092 & 0.499595 & 0.449572 & 0.471380 & 0.471177 & 0.488892 & 0.472382 \\[1.5ex]
arrhytmia & 0.836199 & \bfseries 0.839771 & 0.808447 & 0.722131 & 0.716635 & 0.715732 & 0.810000 & 0.813000 & 0.810096 & 0.806979 & 0.774808 & 0.755613 & 0.581449 & 0.616541 & 0.587612 & 0.605433 & 0.587533 & 0.788507 & 0.800047 & 0.467891 & 0.600228 & 0.507576 & 0.751185 & 0.776849 & 0.777328 & 0.518704 & 0.706781 \\[1.5ex]
cardio & 0.809043 & 0.900412 & 0.839776 & 0.772895 & 0.762637 & 0.666299 & 0.937000 & \bfseries 0.955000 & 0.852541 & 0.931768 & 0.950577 & 0.814162 & 0.938777 & 0.834084 & 0.670739 & 0.670739 & 0.820328 & 0.545784 & 0.686149 & 0.935234 & 0.774689 & 0.806744 & 0.896454 & 0.947740 & 0.950095 & 0.496966 & 0.814101 \\[1.5ex]
annthyroid & 0.556527 & 0.948366 & \bfseries 0.959160 & 0.952567 & 0.955903 & 0.633854 & 0.903000 & 0.878000 & 0.624088 & 0.827367 & 0.672882 & 0.500000 & 0.513267 & 0.397309 & 0.544121 & 0.513324 & 0.469545 & 0.739194 & 0.755748 & 0.546726 & 0.918901 & 0.509363 & 0.641487 & 0.705994 & 0.674360 & 0.491677 & 0.685874 \\[1.5ex]
breastw & 0.974829 & 0.993290 & 0.985318 & 0.984003 & 0.902677 & 0.844547 & 0.989000 & 0.990000 & 0.985129 & 0.987608 & 0.958941 & 0.984328 & \bfseries 0.994921 & 0.987862 & 0.964539 & 0.934430 & 0.857759 & 0.371000 & 0.975301 & 0.978693 & 0.988281 & 0.988281 & 0.972389 & 0.866450 & 0.893671 & 0.892800 & 0.932540 \\
\bottomrule 
\end{tabular}

}
\end{sidewaystable}

\renewcommand{\arraystretch}{2}

\begin{sidewaystable}
\caption{AUROC performance of 6 CoMadOut (CMO*) variants on the left compared to 20 other anomaly detection algorithms on the right on 21 real world datasets. The results are based on top 25\% of all possible principal components.}
\scalebox{0.422}{
\centering
\label{tab:Fig9roc025}
\begin{tabular}{lrrrrrrrrrrrrrrrrrrrrrrrrrrr}
\toprule 
 & CMO & CMO+ & CMO+k & CMO+e & CMO+ke & CMOEns & \shortstack{PCA-\\[1.5ex]MAD++} & \shortstack{PCA-\\[1.5ex]MAD} & HBOS & IF & PCA & PCA(NM) & PCA(r) & PCA(k) & PCA(e) & PCA(ke) & PCA(Ens) & LOF & KNN & OCSVM & MCD & EllipticEnv & MLE & AE & VAE & DeepSVDD & AVG \\[1.5ex]
\midrule 
20news & 0.474693 & 0.459996 & 0.459971 & 0.461755 & 0.461962 & 0.496761 & 0.473976 & 0.465938 & 0.463108 & 0.457231 & 0.471243 & 0.475269 & 0.480010 & 0.473455 & 0.473166 & 0.467730 & 0.487368 & 0.494002 & 0.472295 & 0.468315 & 0.443238 & \bfseries 0.499595 & 0.449572 & 0.471380 & 0.471177 & 0.488892 & 0.471619 \\[1.5ex]
arrhytmia & 0.744583 & 0.808290 & 0.819516 & 0.814885 & \bfseries 0.819987 & 0.618229 & 0.810000 & 0.813000 & 0.810096 & 0.806979 & 0.772217 & 0.739755 & 0.551107 & 0.536112 & 0.549223 & 0.554757 & 0.617562 & 0.788507 & 0.800047 & 0.467891 & 0.611611 & 0.507576 & 0.751185 & 0.776849 & 0.777328 & 0.518704 & 0.699461 \\[1.5ex]
cardio & 0.730059 & 0.855417 & 0.864845 & 0.757858 & 0.751881 & 0.649279 & 0.937000 & \bfseries 0.955000 & 0.852541 & 0.931768 & 0.943257 & 0.666356 & 0.954561 & 0.758046 & 0.895444 & 0.694210 & 0.840925 & 0.545784 & 0.686149 & 0.935234 & 0.772217 & 0.793139 & 0.896454 & 0.947740 & 0.950095 & 0.496966 & 0.810086 \\[1.5ex]
annthyroid & 0.591156 & 0.825793 & 0.820033 & 0.817912 & 0.810979 & 0.636135 & 0.903000 & 0.878000 & 0.624088 & 0.827367 & 0.766821 & 0.500000 & 0.351650 & 0.205853 & 0.398980 & 0.205745 & 0.487904 & 0.739194 & 0.755748 & 0.546726 & \bfseries 0.918901 & 0.509363 & 0.641487 & 0.705994 & 0.674360 & 0.491677 & 0.639802 \\[1.5ex]
breastw & 0.747211 & 0.906795 & 0.897042 & 0.864502 & 0.821973 & 0.770039 & 0.989000 & 0.990000 & 0.985129 & 0.987608 & 0.680218 & 0.990798 & \bfseries 0.996344 & 0.995910 & 0.954597 & 0.940999 & 0.834582 & 0.371000 & 0.975301 & 0.978693 & 0.988281 & 0.988281 & 0.972389 & 0.866450 & 0.893671 & 0.892800 & 0.895370 \\
\bottomrule 
\end{tabular}

}
\end{sidewaystable}

\renewcommand{\arraystretch}{2}

\begin{sidewaystable}
\caption{AUPRC performance of 6 CoMadOut (CMO*) variants on the left side compared to 20 other anomaly detection algorithms on the right side on 21 real world datasets. The results are based on 100\% of all possible principal components.}
\scalebox{0.42}{
\centering
\label{tab:Fig10recall0999}
\begin{tabular}{lrrrrrrrrrrrrrrrrrrrrrrrrrrr}
\toprule 
 & CMO & CMO+ & CMO+k & CMO+e & CMO+ke & CMOEns & \shortstack{PCA-\\[1.5ex]MAD++} & \shortstack{PCA-\\[1.5ex]MAD} & HBOS & IF & PCA & PCA(NM) & PCA(r) & PCA(k) & PCA(e) & PCA(ke) & PCA(Ens) & LOF & KNN & OCSVM & MCD & EllipticEnv & MLE & AE & VAE & DeepSVDD & AVG \\[1.5ex]
\midrule 
20news & 0.043014 & 0.043728 & 0.043743 & 0.045022 & 0.044714 & 0.025000 & 0.044202 & 0.044009 & 0.045596 & 0.044520 & 0.045170 & 0.044633 & 0.045674 & 0.045067 & 0.051484 & 0.050428 & \bfseries 0.094478 & 0.047629 & 0.045746 & 0.044235 & 0.045472 & 0.025000 & 0.044433 & 0.045180 & 0.045261 & 0.046668 & 0.045773 \\[1.5ex]
arrhytmia & \bfseries 0.579657 & 0.559320 & 0.393828 & 0.427435 & 0.432728 & 0.530913 & 0.446137 & 0.482766 & 0.412216 & 0.466397 & 0.387732 & 0.433469 & 0.340079 & 0.337570 & 0.201029 & 0.212896 & 0.366998 & 0.408620 & 0.465353 & 0.106026 & 0.223060 & 0.579478 & 0.278484 & 0.390527 & 0.390427 & 0.165030 & 0.385314 \\[1.5ex]
cardio & 0.248468 & 0.374300 & 0.263532 & 0.395984 & 0.392397 & 0.410154 & \bfseries 0.676025 & 0.565135 & 0.465104 & 0.561869 & 0.610221 & 0.337528 & 0.572526 & 0.341470 & 0.216151 & 0.216151 & 0.582957 & 0.151857 & 0.280918 & 0.529154 & 0.334369 & 0.352234 & 0.461651 & 0.624309 & 0.608646 & 0.110877 & 0.410923 \\[1.5ex]
annthyroid & 0.077672 & 0.625516 & \bfseries 0.662258 & 0.634926 & 0.642141 & 0.560388 & 0.183366 & 0.152975 & 0.181513 & 0.313903 & 0.189672 & 0.537083 & 0.074792 & 0.056518 & 0.077099 & 0.069802 & 0.170754 & 0.204203 & 0.217646 & 0.104409 & 0.504272 & 0.545752 & 0.155945 & 0.229810 & 0.189929 & 0.104233 & 0.287176 \\[1.5ex]
breastw & 0.908464 & 0.986987 & 0.961163 & 0.966615 & 0.862912 & 0.882202 & 0.910473 & 0.956936 & 0.959214 & 0.971922 & 0.957703 & 0.973066 & \bfseries 0.990652 & 0.975218 & 0.935036 & 0.885031 & 0.842015 & 0.270793 & 0.931069 & 0.940113 & 0.968835 & 0.968835 & 0.939205 & 0.846953 & 0.898613 & 0.832505 & 0.904713 \\
\bottomrule 
\end{tabular}

}
\end{sidewaystable}

\renewcommand{\arraystretch}{2}

\begin{sidewaystable}
\caption{AUPRC performance of 6 CoMadOut (CMO*) variants on the left compared to 20 other anomaly detection algorithms on the right on 21 real world datasets. The results are based on top 25\% of all possible principal components.}
\scalebox{0.42}{
\centering
\label{tab:Fig11recall025}
\begin{tabular}{lrrrrrrrrrrrrrrrrrrrrrrrrrrr}
\toprule 
 & CMO & CMO+ & CMO+k & CMO+e & CMO+ke & CMOEns & \shortstack{PCA-\\[1.5ex]MAD++} & \shortstack{PCA-\\[1.5ex]MAD} & HBOS & IF & PCA & PCA(NM) & PCA(r) & PCA(k) & PCA(e) & PCA(ke) & PCA(Ens) & LOF & KNN & OCSVM & MCD & EllipticEnv & MLE & AE & VAE & DeepSVDD & AVG \\[1.5ex]
\midrule 
20news & 0.045046 & 0.045089 & 0.045055 & 0.045210 & 0.045207 & 0.025000 & 0.045700 & 0.044013 & 0.045596 & 0.044520 & 0.045202 & 0.052020 & 0.046471 & 0.044857 & 0.045037 & 0.044865 & \bfseries 0.127271 & 0.047629 & 0.045746 & 0.044235 & 0.044075 & 0.025000 & 0.044433 & 0.045180 & 0.045261 & 0.046668 & 0.047092 \\[1.5ex]
arrhytmia & 0.413201 & 0.493734 & 0.502845 & 0.468550 & 0.471982 & 0.470711 & 0.427767 & 0.482134 & 0.412216 & 0.466397 & 0.385279 & 0.379699 & 0.251167 & 0.251771 & 0.253362 & 0.240960 & 0.392770 & 0.408620 & 0.465353 & 0.106026 & 0.230324 & \bfseries 0.579478 & 0.278484 & 0.390527 & 0.390427 & 0.165030 & 0.376108 \\[1.5ex]
cardio & 0.189214 & 0.340774 & 0.378754 & 0.202116 & 0.196592 & 0.383654 & 0.689035 & 0.658287 & 0.465104 & 0.561869 & 0.588928 & 0.311329 & \bfseries 0.702456 & 0.300773 & 0.533932 & 0.261206 & 0.639284 & 0.151857 & 0.280918 & 0.529154 & 0.382910 & 0.340916 & 0.461651 & 0.624309 & 0.608646 & 0.110877 & 0.419021 \\[1.5ex]
annthyroid & 0.092033 & 0.446492 & 0.439792 & 0.437268 & 0.429431 & 0.509612 & 0.082854 & 0.099994 & 0.181513 & 0.313903 & 0.271237 & 0.537083 & 0.052299 & 0.042979 & 0.056942 & 0.042985 & 0.187291 & 0.204203 & 0.217646 & 0.104409 & 0.504272 & \bfseries 0.545752 & 0.155945 & 0.229810 & 0.189929 & 0.104233 & 0.249227 \\[1.5ex]
breastw & 0.746257 & 0.893161 & 0.870932 & 0.844995 & 0.794632 & 0.820890 & 0.876406 & 0.933908 & 0.959214 & 0.971922 & 0.708315 & 0.983177 & \bfseries 0.992949 & 0.992193 & 0.932391 & 0.914333 & 0.830553 & 0.270793 & 0.931069 & 0.940113 & 0.968835 & 0.968835 & 0.939205 & 0.846953 & 0.898613 & 0.832505 & 0.871659 \\
\bottomrule 
\end{tabular}

}
\end{sidewaystable}

\renewcommand{\arraystretch}{2}

\begin{sidewaystable}
\caption{Precision@n (P@n) performance of 6 CoMadOut (CMO*) variants on the left compared to 20 other anomaly detection algorithms on the right on 21 real world datasets. The results are based on 100\% of all possible principal components.}
\scalebox{0.43}{
\centering
\label{tab:Fig7precn0999}
\begin{tabular}{lrrrrrrrrrrrrrrrrrrrrrrrrrrr}
\toprule 
 & CMO & CMO+ & CMO+k & CMO+e & CMO+ke & CMOEns & \shortstack{PCA-\\[1.5ex]MAD++} & \shortstack{PCA-\\[1.5ex]MAD} & HBOS & IF & PCA & PCA(NM) & PCA(r) & PCA(k) & PCA(e) & PCA(ke) & PCA(Ens) & LOF & KNN & OCSVM & MCD & EllipticEnv & MLE & AE & VAE & DeepSVDD & AVG \\[1.5ex]
\midrule 
20news & 0.030769 & 0.047619 & 0.047619 & 0.051282 & \bfseries 0.054054 & 0.000000 & 0.046154 & 0.030769 & 0.046154 & 0.047685 & 0.030769 & 0.030769 & 0.030769 & 0.030769 & 0.046154 & 0.030769 & 0.000000 & 0.046154 & 0.046154 & 0.030769 & 0.030769 & 0.000000 & 0.027692 & 0.030769 & 0.030769 & 0.030769 & 0.033690 \\[1.5ex]
arrhytmia & 0.530303 & 0.515152 & 0.409091 & 0.424242 & 0.424242 & 0.000000 & 0.503000 & 0.507000 & 0.500000 & 0.481818 & 0.424242 & 0.439394 & 0.348485 & 0.333333 & 0.242424 & 0.242424 & 0.380952 & 0.424242 & 0.469697 & 0.000000 & 0.324242 & \bfseries 1.000000 & 0.375758 & 0.424242 & 0.424242 & 0.157576 & 0.396389 \\[1.5ex]
cardio & 0.250000 & 0.477273 & 0.250000 & 0.409091 & 0.403409 & 0.000000 & 0.579000 & \bfseries 0.649000 & 0.454545 & 0.515909 & 0.607955 & 0.375000 & 0.590909 & 0.375000 & 0.323864 & 0.323864 & 0.000000 & 0.170455 & 0.323864 & 0.505682 & 0.456818 & 0.481818 & 0.454545 & 0.636364 & 0.607955 & 0.143827 & 0.398698 \\[1.5ex]
annthyroid & 0.035581 & 0.559925 & 0.599251 & 0.576779 & 0.573034 & 0.793478 & 0.520000 & 0.451000 & 0.262264 & 0.327341 & 0.237828 & 0.000000 & 0.069288 & 0.028090 & 0.056180 & 0.039326 & 0.000000 & 0.292135 & 0.290262 & 0.112360 & 0.458801 & \bfseries 1.000000 & 0.204120 & 0.275281 & 0.237828 & 0.107491 & 0.311832 \\[1.5ex]
breastw & 0.912134 & 0.945607 & 0.924686 & 0.916318 & 0.775424 & \bfseries 0.954802 & 0.933000 & 0.933000 & 0.936975 & 0.922176 & 0.928870 & 0.928870 & 0.949791 & 0.941423 & 0.878661 & 0.803347 & 0.000000 & 0.129707 & 0.907950 & 0.912134 & 0.938912 & 0.938912 & 0.903766 & 0.766783 & 0.829289 & 0.808939 & 0.835441 \\
\bottomrule 
\end{tabular}

}
\end{sidewaystable}

\renewcommand{\arraystretch}{2}

\begin{sidewaystable}
\caption{Precision@n (P@n) performance of 6 CoMadOut (CMO*) variants on the left compared to 20 other anomaly detection algorithms on the right on 21 real world datasets. The results are based on top 25\% of all possible principal components.}
\scalebox{0.425}{
\centering
\label{tab:Fig12precn025}
\begin{tabular}{lrrrrrrrrrrrrrrrrrrrrrrrrrrr}
\toprule 
 & CMO & CMO+ & CMO+k & CMO+e & CMO+ke & CMOEns & \shortstack{PCA-\\[1.5ex]MAD++} & \shortstack{PCA-\\[1.5ex]MAD} & HBOS & IF & PCA & PCA(NM) & PCA(r) & PCA(k) & PCA(e) & PCA(ke) & PCA(Ens) & LOF & KNN & OCSVM & MCD & EllipticEnv & MLE & AE & VAE & DeepSVDD & AVG \\[1.5ex]
\midrule 
20news & 0.030769 & 0.061171 & \bfseries 0.061929 & 0.060319 & 0.058993 & 0.000000 & 0.030769 & 0.030769 & 0.046154 & 0.047685 & 0.030769 & 0.060330 & 0.043375 & 0.036923 & 0.022462 & 0.020085 & 0.000000 & 0.046154 & 0.046154 & 0.030769 & 0.033846 & 0.000000 & 0.027692 & 0.030769 & 0.030769 & 0.030769 & 0.035362 \\[1.5ex]
arrhytmia & 0.424242 & 0.469697 & 0.500000 & 0.439394 & 0.424242 & 0.562500 & 0.503000 & 0.507000 & 0.500000 & 0.481818 & 0.424242 & 0.378788 & 0.242424 & 0.242424 & 0.257576 & 0.257576 & 0.000000 & 0.424242 & 0.469697 & 0.000000 & 0.330303 & \bfseries 1.000000 & 0.375758 & 0.424242 & 0.424242 & 0.157576 & 0.393115 \\[1.5ex]
cardio & 0.221591 & 0.363636 & 0.392045 & 0.232955 & 0.221591 & 0.000000 & 0.579000 & 0.649000 & 0.454545 & 0.515909 & 0.590909 & 0.357955 & \bfseries 0.664773 & 0.278409 & 0.556818 & 0.261364 & 0.000000 & 0.170455 & 0.323864 & 0.505682 & 0.476136 & 0.464773 & 0.454545 & 0.636364 & 0.607955 & 0.143827 & 0.389388 \\[1.5ex]
annthyroid & 0.063670 & 0.419476 & 0.419476 & 0.411985 & 0.406367 & 0.683258 & 0.520000 & 0.451000 & 0.262264 & 0.327341 & 0.307116 & 0.000000 & 0.022472 & 0.003745 & 0.031835 & 0.001873 & 0.000000 & 0.292135 & 0.290262 & 0.112360 & 0.458801 & \bfseries 1.000000 & 0.204120 & 0.275281 & 0.237828 & 0.107491 & 0.281160 \\[1.5ex]
breastw & 0.679612 & 0.903930 & 0.877119 & 0.777293 & 0.708696 & 0.924658 & 0.933000 & 0.933000 & 0.936975 & 0.922176 & 0.592437 & 0.941423 & \bfseries 0.958159 & \bfseries 0.958159 & 0.866109 & 0.832636 & 0.816143 & 0.129707 & 0.907950 & 0.912134 & 0.938912 & 0.938912 & 0.903766 & 0.766783 & 0.829289 & 0.808939 & 0.834535 \\
\bottomrule 
\end{tabular}

}
\end{sidewaystable}

\renewcommand{\arraystretch}{2}

\begin{sidewaystable}
\caption{Runtime in seconds of 6 CoMadOut (CMO*) variants on the left side compared to 20 other anomaly detection algorithms on the right side on 21 real world datasets. The results are based on 100\% of all possible principal components.}
\scalebox{0.41}{
\centering
\label{tab:Fig8runtime0999}
\begin{tabular}{lrrrrrrrrrrrrrrrrrrrrrrrrrrr}
\toprule 
 & CMO & CMO+ & CMO+k & CMO+e & CMO+ke & CMOEns & \shortstack{PCA-\\[1.5ex]MAD++} & \shortstack{PCA-\\[1.5ex]MAD} & HBOS & IF & PCA & PCA(NM) & PCA(r) & PCA(k) & PCA(e) & PCA(ke) & PCA(Ens) & LOF & KNN & OCSVM & MCD & EllipticEnv & MLE & AE & VAE & DeepSVDD & AVG \\[1.5ex]
\midrule 
20news & 215.852400 & 77.197400 & 77.197400 & 77.197400 & 77.197400 & 77.197400 & 0.672800 & 0.624600 & \bfseries 0.367400 & 1.252400 & 0.782400 & 186.229600 & 40.356600 & 39.902800 & 40.512400 & 40.419200 & 40.149600 & 2.989200 & 2.997600 & 0.871000 & 145.237800 & 148.041400 & 0.395600 & 8.374400 & 10.467200 & 1.851000 & 50.551323 \\[1.5ex]
arrhytmia & 12.865400 & 7.713000 & 7.713000 & 7.713000 & 7.713000 & 7.713000 & 0.171400 & 0.167600 & \bfseries 0.039800 & 0.284000 & 0.052600 & 8.344800 & 3.025400 & 3.081600 & 3.004200 & 3.031200 & 3.032600 & 0.129800 & 0.151000 & 0.079000 & 4.984800 & 3.561400 & 0.170800 & 3.161200 & 4.836400 & 1.179000 & 3.612269 \\[1.5ex]
cardio & 1.609600 & 2.896200 & 2.896200 & 2.896200 & 2.896200 & 2.896200 & 0.026800 & 0.016200 & \bfseries 0.005000 & 0.284200 & 0.006600 & 0.422200 & 1.594400 & 1.584400 & 1.577200 & 1.585800 & 1.579400 & 0.124600 & 0.197800 & 0.237000 & 0.595200 & 0.645800 & 0.008600 & 8.916200 & 15.061800 & 1.905200 & 2.017885 \\[1.5ex]
annthyroid & 1.486000 & 2.142800 & 2.142800 & 2.142800 & 2.142800 & 2.142800 & 0.009200 & 0.007400 & \bfseries 0.003600 & 0.524400 & 0.005200 & 0.328600 & 0.915200 & 0.903000 & 0.897600 & 0.885600 & 0.903600 & 0.453600 & 0.651800 & 3.005600 & 1.285400 & 1.126600 & 0.148000 & 21.275000 & 36.207200 & 4.616400 & 3.321269 \\[1.5ex]
breastw & 1.165400 & 2.324000 & 2.324000 & 2.324000 & 2.324000 & 2.324000 & 0.006200 & 0.005000 & 0.002600 & 0.201200 & \bfseries 0.002000 & 0.061400 & 1.059800 & 1.044200 & 1.045000 & 1.019800 & 1.039000 & 0.013200 & 0.039800 & 0.029800 & 0.353200 & 0.331000 & 0.005400 & 3.998200 & 6.059800 & 1.335800 & 1.170685 \\
\bottomrule 
\end{tabular}

}
\end{sidewaystable}

\renewcommand{\arraystretch}{2}

\begin{sidewaystable}
\caption{Runtime in seconds of 6 CoMadOut (CMO*) variants on the left compared to 20 other anomaly detection algorithms on the right on 21 real world datasets. The results are based on top 25\% of all possible principal components.}
\scalebox{0.415}{
\centering
\label{tab:Fig13runtime025}
\begin{tabular}{lrrrrrrrrrrrrrrrrrrrrrrrrrrr}
\toprule 
 & CMO & CMO+ & CMO+k & CMO+e & CMO+ke & CMOEns & \shortstack{PCA-\\[1.5ex]MAD++} & \shortstack{PCA-\\[1.5ex]MAD} & HBOS & IF & PCA & PCA(NM) & PCA(r) & PCA(k) & PCA(e) & PCA(ke) & PCA(Ens) & LOF & KNN & OCSVM & MCD & EllipticEnv & MLE & AE & VAE & DeepSVDD & AVG \\[1.5ex]
\midrule 
20news & 82.176400 & 31.346000 & 31.346000 & 31.346000 & 31.346000 & 31.346000 & 1.283000 & 1.587800 & 0.399000 & 1.253000 & 0.289600 & 10.209600 & 3.466800 & 3.328200 & 3.388800 & 3.436600 & 3.463000 & 3.050800 & 3.017800 & 0.892800 & 84.104000 & 85.545400 & \bfseries 0.168000 & 7.382600 & 10.389800 & 1.890400 & 17.978977 \\[1.5ex]
arrhytmia & 5.270600 & 6.257800 & 6.257800 & 6.257800 & 6.257800 & 6.257800 & 0.053800 & 0.060600 & 0.040200 & 0.282000 & 0.031800 & 0.640400 & 1.512000 & 1.486200 & 1.472000 & 1.493600 & 1.500800 & 0.134000 & 0.154800 & 0.087600 & 1.008200 & 0.903600 & \bfseries 0.022200 & 3.148800 & 4.418800 & 1.197600 & 2.161869 \\[1.5ex]
cardio & 1.155200 & 1.944000 & 1.944000 & 1.944000 & 1.944000 & 1.944000 & 0.012800 & 0.010000 & 0.005000 & 0.282800 & \bfseries 0.004200 & 0.095800 & 0.799800 & 0.776200 & 0.783600 & 0.788600 & 0.780800 & 0.127600 & 0.182800 & 0.241000 & 0.445800 & 0.441000 & 0.014400 & 8.874600 & 13.161000 & 1.933800 & 1.562954 \\[1.5ex]
annthyroid & 1.158400 & 1.721800 & 1.721800 & 1.721800 & 1.721800 & 1.721800 & 0.011600 & 0.008200 & \bfseries 0.003800 & 0.511600 & 0.004000 & 0.159200 & 0.595800 & 0.574200 & 0.568400 & 0.567200 & 0.573000 & 0.456200 & 0.601600 & 3.075800 & 1.263200 & 1.139600 & 0.141800 & 20.896600 & 35.776000 & 4.598000 & 3.126662 \\[1.5ex]
breastw & 0.978600 & 1.734200 & 1.734200 & 1.734200 & 1.734200 & 1.734200 & 0.006800 & 0.007400 & 0.003000 & 0.200600 & \bfseries 0.001600 & 0.023800 & 0.619000 & 0.594200 & 0.589000 & 0.600800 & 0.590800 & 0.013600 & 0.035400 & 0.029200 & 0.344200 & 0.337000 & 0.003000 & 4.074800 & 5.978800 & 1.334600 & 0.962969 \\
\bottomrule 
\end{tabular}

}
\end{sidewaystable}

\renewcommand{\arraystretch}{2}

\begin{sidewaystable}
\caption{AUROC of 6 CoMadOut (CMO*) variants on the left compared to 20 other anomaly detection algorithms on the right on 21 real world datasets. The results are based on parameter tuning.}
\scalebox{0.415}{
\centering
\label{tab:FigPTauroc}
\begin{tabular}{lrrrrrrrrrrrrrrrrrrrrrrrrrrr}
\toprule 
 & CMO & CMO+ & CMO+k & CMO+e & CMO+ke & CMOEns & \shortstack{PCA-\\[1.5ex]MAD++} & \shortstack{PCA-\\[1.5ex]MAD} & HBOS & IF & PCA & PCA(NM) & PCA(r) & PCA(k) & PCA(e) & PCA(ke) & PCA(Ens) & LOF & KNN & OCSVM & MCD & EllipticEnv & MLE & AE & VAE & DeepSVDD & AVG \\[1.5ex]
\midrule 
20news & 0.475353 & 0.461875 & 0.461514 & 0.463058 & 0.462024 & 0.496761 & 0.473976 & 0.465856 & 0.463108 & 0.466087 & 0.471878 & 0.475269 & 0.480010 & 0.472949 & 0.532370 & 0.525842 & 0.493846 & \bfseries 0.551106 & 0.472295 & 0.488546 & 0.486484 & 0.499595 & 0.449572 & 0.471327 & 0.469378 & 0.516463 & 0.482559 \\[1.5ex]
arrhytmia & 0.836199 & \bfseries 0.839771 & 0.819516 & 0.814885 & 0.819987 & 0.715732 & 0.773238 & 0.806799 & 0.810096 & 0.801727 & 0.775318 & 0.758165 & 0.581449 & 0.609279 & 0.632517 & 0.681700 & 0.629416 & 0.814100 & 0.816337 & 0.578800 & 0.616741 & 0.507576 & 0.751185 & 0.787690 & 0.787745 & 0.520427 & 0.726400 \\[1.5ex]
cardio & 0.844565 & 0.900412 & 0.864845 & 0.772895 & 0.762637 & 0.666299 & 0.946879 & 0.953385 & 0.852541 & 0.932072 & 0.950577 & 0.814162 & 0.954561 & 0.834084 & 0.895444 & 0.819888 & 0.840925 & 0.852142 & 0.909922 & 0.943099 & 0.848579 & 0.881495 & 0.896454 & \bfseries 0.958090 & 0.957505 & 0.488391 & 0.859302 \\[1.5ex]
annthyroid & 0.591156 & 0.950817 & 0.963967 & 0.960899 & \bfseries 0.971759 & 0.635015 & 0.667287 & 0.591531 & 0.898074 & 0.822356 & 0.766821 & 0.494599 & 0.513226 & 0.397309 & 0.550905 & 0.487545 & 0.495182 & 0.706858 & 0.724567 & 0.555202 & 0.931701 & 0.722454 & 0.641487 & 0.708743 & 0.692981 & 0.499053 & 0.690057 \\[1.5ex]
breastw & 0.977025 & 0.993290 & 0.985318 & 0.984003 & 0.902682 & 0.844547 & 0.947670 & 0.963681 & 0.985129 & 0.986924 & 0.960195 & 0.990798 & \bfseries 0.996344 & 0.995910 & 0.981030 & 0.971512 & 0.880918 & 0.549917 & 0.992739 & 0.981803 & 0.991080 & 0.991080 & 0.972389 & 0.847343 & 0.803362 & 0.953185 & 0.939611 \\
\bottomrule 
\end{tabular}

}
\end{sidewaystable}

\renewcommand{\arraystretch}{2}

\begin{sidewaystable}
\caption{Average Precision of 6 CoMadOut (CMO*) variants on the left compared to 20 other anomaly detection algorithms on the right on 21 real world datasets. The results are based on parameter tuning.}
\scalebox{0.415}{
\centering
\label{tab:FigPTap}
\begin{tabular}{lrrrrrrrrrrrrrrrrrrrrrrrrrrr}
\toprule 
 & CMO & CMO+ & CMO+k & CMO+e & CMO+ke & CMOEns & \shortstack{PCA-\\[1.5ex]MAD++} & \shortstack{PCA-\\[1.5ex]MAD} & HBOS & IF & PCA & PCA(NM) & PCA(r) & PCA(k) & PCA(e) & PCA(ke) & PCA(Ens) & LOF & KNN & OCSVM & MCD & EllipticEnv & MLE & AE & VAE & DeepSVDD & AVG \\[1.5ex]
\midrule 
20news & 0.046851 & 0.046912 & 0.046862 & 0.047042 & 0.046895 & 0.050000 & 0.047184 & 0.045716 & 0.047568 & 0.046621 & 0.046820 & 0.053226 & 0.048010 & 0.046875 & 0.053103 & 0.052243 & 0.049470 & \bfseries 0.058407 & 0.047434 & 0.047527 & 0.048946 & 0.050000 & 0.045824 & 0.046795 & 0.046742 & 0.054746 & 0.048762 \\[1.5ex]
arrhytmia & \bfseries 0.583627 & 0.562570 & 0.507573 & 0.473604 & 0.477107 & 0.311826 & 0.456998 & 0.489558 & 0.419830 & 0.461250 & 0.395590 & 0.441898 & 0.343874 & 0.347989 & 0.344357 & 0.399053 & 0.216230 & 0.505823 & 0.516195 & 0.190999 & 0.271904 & 0.158957 & 0.285995 & 0.409948 & 0.410870 & 0.169393 & 0.390501 \\[1.5ex]
cardio & 0.363950 & 0.377497 & 0.380866 & 0.397465 & 0.393873 & 0.201126 & 0.699126 & 0.659719 & 0.466654 & 0.568205 & 0.612072 & 0.339740 & \bfseries 0.703615 & 0.343690 & 0.535627 & 0.318869 & 0.397577 & 0.297684 & 0.511820 & 0.575303 & 0.456945 & 0.480982 & 0.463834 & 0.658245 & 0.668994 & 0.130333 & 0.461685 \\[1.5ex]
annthyroid & 0.091705 & 0.634651 & 0.682934 & 0.668410 & \bfseries 0.715989 & 0.276092 & 0.184498 & 0.153783 & 0.417227 & 0.302148 & 0.272873 & 0.074167 & 0.075477 & 0.056714 & 0.085658 & 0.075755 & 0.073521 & 0.254906 & 0.222985 & 0.108630 & 0.544366 & 0.291855 & 0.157324 & 0.242294 & 0.220264 & 0.082977 & 0.267969 \\[1.5ex]
breastw & 0.922811 & 0.987017 & 0.961325 & 0.966530 & 0.863497 & 0.777642 & 0.928407 & 0.957041 & 0.959415 & 0.970394 & 0.963820 & 0.982058 & \bfseries 0.992965 & 0.992210 & 0.960092 & 0.936540 & 0.817800 & 0.364565 & 0.985035 & 0.948286 & 0.981351 & 0.981351 & 0.939455 & 0.824482 & 0.819969 & 0.941720 & 0.912530 \\
\bottomrule 
\end{tabular}

}
\end{sidewaystable}

\renewcommand{\arraystretch}{2}

\begin{sidewaystable}
\caption{AUPRC of 6 CoMadOut (CMO*) variants on the left compared to 20 other anomaly detection algorithms on the right on 21 real world datasets. The results are based on parameter tuning.}
\scalebox{0.415}{
\centering
\label{tab:FigPTauprc}
\begin{tabular}{lrrrrrrrrrrrrrrrrrrrrrrrrrrr}
\toprule 
 & CMO & CMO+ & CMO+k & CMO+e & CMO+ke & CMOEns & \shortstack{PCA-\\[1.5ex]MAD++} & \shortstack{PCA-\\[1.5ex]MAD} & HBOS & IF & PCA & PCA(NM) & PCA(r) & PCA(k) & PCA(e) & PCA(ke) & PCA(Ens) & LOF & KNN & OCSVM & MCD & EllipticEnv & MLE & AE & VAE & DeepSVDD & AVG \\[1.5ex]
\midrule 
20news & 0.045087 & 0.045445 & 0.045435 & 0.045531 & 0.045396 & 0.025000 & 0.045700 & 0.044014 & 0.045596 & 0.045158 & 0.045200 & 0.052020 & 0.046471 & 0.045067 & 0.051484 & 0.050428 & \bfseries 0.125791 & 0.056209 & 0.045746 & 0.046002 & 0.047466 & 0.025000 & 0.044433 & 0.045175 & 0.045226 & 0.052387 & 0.048326 \\[1.5ex]
arrhytmia & \bfseries 0.579657 & 0.559320 & 0.502845 & 0.468550 & 0.471982 & 0.530913 & 0.446142 & 0.482766 & 0.412216 & 0.454756 & 0.388234 & 0.435317 & 0.340079 & 0.342783 & 0.338944 & 0.393997 & 0.406511 & 0.499126 & 0.509620 & 0.185030 & 0.260442 & 0.579478 & 0.278484 & 0.402600 & 0.403544 & 0.164022 & 0.416821 \\[1.5ex]
cardio & 0.361422 & 0.374300 & 0.378754 & 0.395984 & 0.392397 & 0.410154 & 0.698104 & 0.658287 & 0.465104 & 0.566707 & 0.610221 & 0.337528 & \bfseries 0.702456 & 0.341470 & 0.533932 & 0.316701 & 0.639284 & 0.295093 & 0.509565 & 0.570816 & 0.453898 & 0.477959 & 0.461651 & 0.656460 & 0.667306 & 0.128688 & 0.477086 \\[1.5ex]
annthyroid & 0.092033 & 0.634254 & 0.682550 & 0.668018 & \bfseries 0.715613 & 0.568362 & 0.183366 & 0.153177 & 0.415410 & 0.301012 & 0.271237 & 0.037083 & 0.074993 & 0.056518 & 0.084714 & 0.074856 & 0.239711 & 0.254075 & 0.221379 & 0.107173 & 0.543865 & 0.456563 & 0.155945 & 0.240684 & 0.218641 & 0.082183 & 0.289747 \\[1.5ex]
breastw & 0.920430 & 0.986987 & 0.961163 & 0.966615 & 0.862921 & 0.882202 & 0.928152 & 0.956936 & 0.959214 & 0.970304 & 0.963770 & 0.983177 & \bfseries 0.992949 & 0.992193 & 0.959983 & 0.936316 & 0.903003 & 0.361328 & 0.985234 & 0.948061 & 0.981304 & 0.981304 & 0.939205 & 0.824005 & 0.819716 & 0.941596 & 0.919541 \\
\bottomrule 
\end{tabular}

}
\end{sidewaystable}

\renewcommand{\arraystretch}{2}

\begin{sidewaystable}
\caption{P@N of 6 CoMadOut (CMO*) variants on the left compared to 20 other anomaly detection algorithms on the right on 21 real world datasets. The results are based on parameter tuning.}
\scalebox{0.415}{
\centering
\label{tab:FigPTprecn}
\begin{tabular}{lrrrrrrrrrrrrrrrrrrrrrrrrrrr}
\toprule 
 & CMO & CMO+ & CMO+k & CMO+e & CMO+ke & CMOEns & \shortstack{PCA-\\[1.5ex]MAD++} & \shortstack{PCA-\\[1.5ex]MAD} & HBOS & IF & PCA & PCA(NM) & PCA(r) & PCA(k) & PCA(e) & PCA(ke) & PCA(Ens) & LOF & KNN & OCSVM & MCD & EllipticEnv & MLE & AE & VAE & DeepSVDD & AVG \\[1.5ex]
\midrule 
20news & 0.030769 & 0.060606 & 0.062500 & 0.058824 & 0.057143 & 0.000000 & 0.030769 & 0.030769 & 0.046154 & 0.036923 & 0.030769 & 0.060330 & 0.043375 & 0.030769 & 0.046154 & 0.030769 & 0.000000 & \bfseries 0.076923 & 0.046154 & 0.030769 & 0.046154 & 0.000000 & 0.027692 & 0.030769 & 0.030769 & 0.058462 & 0.038628 \\[1.5ex]
arrhytmia & 0.530303 & 0.515152 & 0.500000 & 0.439394 & 0.424242 & 0.000000 & 0.484848 & 0.469697 & 0.500000 & 0.475758 & 0.424242 & 0.454545 & 0.348485 & 0.303030 & 0.348485 & 0.363636 & 0.000000 & 0.530303 & 0.515152 & 0.169231 & 0.354545 & \bfseries 1.000000 & 0.375758 & 0.454545 & 0.454545 & 0.163636 & 0.407674 \\[1.5ex]
cardio & 0.357955 & 0.477273 & 0.392045 & 0.409091 & 0.403409 & 0.000000 & \bfseries 0.687500 & 0.659091 & 0.454545 & 0.513636 & 0.607955 & 0.375000 & 0.664773 & 0.375000 & 0.556818 & 0.329545 & 0.000000 & 0.232955 & 0.511364 & 0.573864 & 0.528409 & 0.551136 & 0.454545 & 0.657955 & 0.664773 & 0.137500 & 0.445236 \\[1.5ex]
annthyroid & 0.063670 & 0.567416 & 0.621723 & 0.612360 & 0.653558 & \bfseries 0.807692 & 0.264045 & 0.164794 & 0.385768 & 0.330288 & 0.307116 & 0.000000 & 0.071161 & 0.028090 & 0.089888 & 0.076779 & 0.000000 & 0.325843 & 0.286517 & 0.112360 & 0.477528 & 0.619101 & 0.204120 & 0.283146 & 0.264045 & 0.074157 & 0.295814 \\[1.5ex]
breastw & 0.916318 & 0.945607 & 0.924686 & 0.916318 & 0.775424 & 0.954802 & 0.912134 & 0.920502 & 0.936975 & 0.921339 & 0.933054 & 0.941423 & \bfseries 0.958159 & \bfseries 0.958159 & 0.912134 & 0.895397 & 0.944724 & 0.330544 & 0.945148 & 0.928870 & 0.928870 & 0.928870 & 0.903766 & 0.737238 & 0.735565 & 0.888703 & 0.884413 \\
\bottomrule 
\end{tabular}

}
\end{sidewaystable}

\renewcommand{\arraystretch}{2}

\begin{sidewaystable}
\caption{Runtime of 6 CoMadOut (CMO*) variants on the left compared to 20 other anomaly detection algorithms on the right on 21 real world datasets. The results are based on parameter tuning.}
\scalebox{0.415}{
\centering
\label{tab:FigPTruntime}
\begin{tabular}{lrrrrrrrrrrrrrrrrrrrrrrrrrrr}
\toprule 
 & CMO & CMO+ & CMO+k & CMO+e & CMO+ke & CMOEns & \shortstack{PCA-\\[1.5ex]MAD++} & \shortstack{PCA-\\[1.5ex]MAD} & HBOS & IF & PCA & PCA(NM) & PCA(r) & PCA(k) & PCA(e) & PCA(ke) & PCA(Ens) & LOF & KNN & OCSVM & MCD & EllipticEnv & MLE & AE & VAE & DeepSVDD & AVG \\[1.5ex]
\midrule 
20news & 199.937000 & 184.200600 & 184.200600 & 184.200600 & 184.200600 & 184.200600 & 1.515000 & 1.456600 & 0.979400 & \bfseries 0.049600 & 0.247400 & 45.448200 & 4.708200 & 41.385200 & 42.129200 & 41.773400 & 25.580000 & 0.554400 & 1.762800 & 0.222800 & 130.954800 & 113.684000 & 0.464200 & 7.479400 & 3.162800 & 3.137800 & 61.062892 \\[1.5ex]
arrhytmia & 36.508000 & 32.706800 & 32.706800 & 32.706800 & 32.706800 & 32.706800 & 0.260400 & 0.265800 & 0.094200 & 0.191200 & \bfseries 0.043000 & 17.815800 & 3.488200 & 2.575400 & 2.552400 & 2.545600 & 2.054400 & 0.136200 & 0.208400 & 0.094800 & 4.837600 & 5.117200 & 0.497600 & 1.684600 & 2.281600 & 1.747600 & 9.559000 \\[1.5ex]
cardio & 2.592400 & 11.937000 & 11.937000 & 11.937000 & 11.937000 & 11.937000 & 0.027600 & 0.022400 & 0.010600 & 0.363400 & \bfseries 0.008400 & 1.304200 & 0.672400 & 1.163200 & 0.607200 & 0.996000 & 0.622800 & 0.334400 & 0.548000 & 0.421800 & 0.904200 & 0.880200 & 0.013600 & 3.101200 & 2.315200 & 1.759000 & 3.013585 \\[1.5ex]
annthyroid & 1.287000 & 7.613800 & 7.613800 & 7.613800 & 7.613800 & 7.613800 & 0.016000 & 0.012000 & 0.007000 & 0.051800 & \bfseries 0.006200 & 1.039800 & 0.724400 & 0.732400 & 0.673000 & 0.670000 & 0.677200 & 0.437200 & 1.205400 & 0.662600 & 2.044600 & 2.549600 & 0.127200 & 4.407200 & 2.841000 & 3.193400 & 2.362846 \\[1.5ex]
breastw & 1.393400 & 9.923000 & 9.923000 & 9.923000 & 9.923000 & 9.923000 & 0.016200 & 0.013000 & 0.005000 & 0.155600 & \bfseries 0.003200 & 0.056600 & 0.547000 & 0.502800 & 0.566200 & 0.571000 & 0.570600 & 0.029400 & 0.125600 & 0.050400 & 0.842800 & 0.778400 & 0.003600 & 2.541000 & 4.261800 & 2.564000 & 2.508177 \\
\bottomrule 
\end{tabular}

}
\end{sidewaystable}

\renewcommand{\arraystretch}{2}

\begin{sidewaystable}
\centering
\caption{Model- and hyperparameters used for dataset specific parameter tuning.}
\scalebox{0.51}{
\label{tab:FigPTparams}
\begin{tabular}{llllllllllllllllllllll}
\toprule
{} & 20news & arrhytmia &  cardio & annthyroid & breastw & letter & thyroid & mammography &    pima &    musk & optdigits & pendigits & mnist & shuttle & satellite & satimage-2 &  wine & vowels & glass &   wbc &  boston \\
param-alg                    &        &           &         &            &         &        &         &             &         &         &           &           &       &         &           &            &       &        &       &       &         \\
\midrule
n\_components-CMO             &    0.5 &       1.0 &     1.0 &       0.25 &     1.0 &   0.25 &     1.0 &        0.75 &     1.0 &     1.0 &       0.5 &       1.0 &   1.0 &     0.5 &      0.75 &        0.5 &   1.0 &    1.0 &   1.0 &  0.25 &     1.0 \\
n\_components-CMO+            &   0.25 &       1.0 &     1.0 &       0.75 &     1.0 &    1.0 &    0.25 &        0.75 &     1.0 &    0.25 &      0.75 &      0.25 &   1.0 &    0.75 &       1.0 &        1.0 &  0.25 &    1.0 &  0.75 &  0.25 &     1.0 \\
n\_components-CMO+k           &   0.25 &      0.25 &    0.25 &       0.75 &     1.0 &    1.0 &    0.25 &        0.75 &     1.0 &    0.25 &       0.5 &      0.25 &   1.0 &     1.0 &       1.0 &        1.0 &  0.25 &    1.0 &   0.5 &  0.25 &     1.0 \\
n\_components-CMO+e           &   0.25 &      0.25 &     1.0 &       0.75 &     1.0 &    1.0 &    0.25 &        0.75 &     1.0 &    0.25 &      0.75 &      0.25 &   1.0 &     1.0 &       0.5 &       0.25 &  0.25 &   0.75 &  0.75 &   1.0 &     0.5 \\
n\_components-CMO+ke          &   0.25 &      0.25 &     1.0 &       0.75 &     1.0 &    0.5 &    0.25 &        0.75 &     1.0 &    0.25 &      0.75 &      0.25 &  0.75 &    0.75 &       0.5 &       0.25 &  0.25 &   0.75 &   0.5 &   1.0 &     0.5 \\
\bottomrule
\end{tabular}

}
\end{sidewaystable}

\end{appendices}

\FloatBarrier

\bibliography{sn-bibliography}

\end{document}